\documentclass[journal]{IEEEtran}

\ifCLASSINFOpdf
  \usepackage[pdftex]{graphicx}
  \graphicspath{{figures/}}
  \DeclareGraphicsExtensions{.pdf,.jpg,.jpeg,.png}
\else
\fi

\usepackage{amsmath}
\usepackage[dvipsnames,cmyk]{xcolor}
\usepackage{algorithm}
\usepackage{algorithmicx}
\usepackage{algpseudocode}
\usepackage{adjustbox}
\usepackage{balance}
\usepackage{array}
\usepackage{csquotes}
\usepackage{blindtext}
\usepackage{booktabs}
\usepackage{textcomp}
\usepackage{gensymb}
\usepackage{multirow}
\usepackage{pdfpages}
\usepackage{hhline}
\usepackage{rotating}
\usepackage{xparse}
\usepackage{pgfplots}
\usepackage{pgfplotstable}
\usepackage{amsmath}
\usepackage[T1]{fontenc}
\usepackage{makecell}
\usepackage{mathtools, nccmath}
\usepackage{nicefrac}
\usepackage{fdsymbol}

\graphicspath{{figures/}}

\usepgfplotslibrary{colorbrewer, statistics}
\usetikzlibrary{pgfplots.groupplots}
\usetikzlibrary{matrix}

\pgfplotsset{
	compat=1.14,
	colormap/Dark2-8,
	colormap/BrBG-8,
	colormap/Set1-8,
	colormap/Spectral-8,
	colormap/Paired-8
}

\newlength{\subsubsectionspace}
\newcommand*{\unit}[1]{\,\textrm{#1}}

\newcolumntype{S}{p{15.0mm}<{\raggedleft}}

\newlength{\condensegraphicsspacing}
\newlength{\jointPositionsPlotHeights}
\newcommand*{\abs}[1]{\| #1 \|}
\renewcommand*{\vec}[1]{\boldsymbol{#1}}
\newcommand*{\mat}[1]{\boldsymbol{#1}}
\newcommand*{\idx}[1]{\text{#1}}
\newcommand*{\func}[1]{\text{#1}}
\newcommand*{\transform}[2]{\boldsymbol{T}^{\mathcal{#1} \rightarrow \mathcal{#2}}}
\newcommand*{\jf}[1]{\mathcal{J}_{#1}}

\newcommand*{\transpose}[1]{#1 ^\text{T} }
\newcommand*{\set}[1]{\{ #1 \}}

\definecolor{Pantone301}{HTML}{005293}
\definecolor{Pantone542}{HTML}{64A0C8}
\definecolor{Pantone283}{HTML}{98C6EA}
\definecolor{Pantone383}{HTML}{A2AD00}
\definecolor{Pantone158}{HTML}{E37222}
\definecolor{Pantone7527}{HTML}{DAD7CB}

\pgfplotscreateplotcyclelist{trajectory-colors}{
	{Pantone7527},{Pantone7527},{Pantone7527},{Pantone7527},{Pantone7527},{Pantone7527},{Pantone7527},
	{Pantone158}, {Pantone158}, {Pantone158}, {Pantone158}, {Pantone158}, {Pantone158}, {Pantone158},
	{Pantone383}, {Pantone383}, {Pantone383}, {Pantone383}, {Pantone383}, {Pantone383}, {Pantone383},
	{Pantone283}, {Pantone283}, {Pantone283}, {Pantone283}, {Pantone283}, {Pantone283}, {Pantone283},
	{Pantone301}, {Pantone301}, {Pantone301}, {Pantone301}, {Pantone301}, {Pantone301}, {Pantone301},
}

\tikzset{
	basiclinestyle/.style = {line width = 0.8pt, solid, line join=round},
}

\pgfplotsset{
	trajectory-linestyles/.style = {
		every axis plot no  1/.append style = {basiclinestyle, color=black, solid},
		every axis plot no  2/.append style = {basiclinestyle, color=black, densely dotted},
		every axis plot no  3/.append style = {basiclinestyle, color=black, densely dashed},
		every axis plot no  4/.append style = {basiclinestyle, color=black, densely dashdotted},
		every axis plot no  5/.append style = {basiclinestyle, color=black, densely dashdotdotted},
		every axis plot no  6/.append style = {basiclinestyle, color=black, dashed},
		every axis plot no  7/.append style = {basiclinestyle, color=black, dashdotted},
		every axis plot no  8/.append style = {basiclinestyle, color=black, solid},
		every axis plot no  9/.append style = {basiclinestyle, color=black, densely dotted},
		every axis plot no 10/.append style = {basiclinestyle, color=black, densely dashed},
		every axis plot no 11/.append style = {basiclinestyle, color=black, densely dashdotted},
		every axis plot no 12/.append style = {basiclinestyle, color=black, densely dashdotdotted},
		every axis plot no 13/.append style = {basiclinestyle, color=black, dashed},
		every axis plot no 14/.append style = {basiclinestyle, color=black, dashdotted},
		every axis plot no 15/.append style = {basiclinestyle, color=black, solid},
		every axis plot no 16/.append style = {basiclinestyle, color=black, densely dotted},
		every axis plot no 17/.append style = {basiclinestyle, color=black, densely dashed},
		every axis plot no 18/.append style = {basiclinestyle, color=black, densely dashdotted},
		every axis plot no 19/.append style = {basiclinestyle, color=black, densely dashdotdotted},
		every axis plot no 20/.append style = {basiclinestyle, color=black, dashed},
		every axis plot no 21/.append style = {basiclinestyle, color=black, dashdotted},
		every axis plot no 22/.append style = {basiclinestyle, color=black, solid},
		every axis plot no 23/.append style = {basiclinestyle, color=black, densely dotted},
		every axis plot no 24/.append style = {basiclinestyle, color=black, densely dashed},
		every axis plot no 25/.append style = {basiclinestyle, color=black, densely dashdotted},
		every axis plot no 26/.append style = {basiclinestyle, color=black, densely dashdotdotted},
		every axis plot no 27/.append style = {basiclinestyle, color=black, dashed},
		every axis plot no 28/.append style = {basiclinestyle, color=black, dashdotted},
		every axis plot no 29/.append style = {basiclinestyle, color=black, solid},
		every axis plot no 30/.append style = {basiclinestyle, color=black, densely dotted},
		every axis plot no 31/.append style = {basiclinestyle, color=black, densely dashed},
		every axis plot no 32/.append style = {basiclinestyle, color=black, densely dashdotted},
		every axis plot no 33/.append style = {basiclinestyle, color=black, densely dashdotdotted},
		every axis plot no 34/.append style = {basiclinestyle, color=black, dashed},
		every axis plot no 35/.append style = {basiclinestyle, color=black, dashdotted},
	}
}

\algnewcommand\algorithmicforeach{\textbf{for each}}
\algdef{S}[FOR]{ForEach}[1]{\algorithmicforeach\ #1\ \algorithmicdo}

\begin{document}%
\title{Robot Self-Calibration Using Actuated 3D Sensors}

\author{Arne Peters$^{1}$%
	\thanks{This manuscript is part of projects that have received funding from the European Union's Horizon 2020 research and innovation programme under grant agreement No 870133.}%
	\thanks{$^{1}$	All research was conducted at the Chair of Robotics, Artificial Intelligence and Real-time Systems, Technical University of Munich (TUM), Boltzmannstr. 3, 85748 Garching, Germany (see http://www6.in.tum.de).
	{\tt\small arne.peters@tum.de}}%
}

\maketitle

\begin{abstract}
Both, robot and hand-eye calibration haven been object to research for decades.
While current approaches manage to precisely and robustly identify the parameters of a robot's kinematic model, they still rely on external devices, such as calibration objects, markers and/or external sensors.
Instead of trying to fit the recorded measurements to a model of a known object, this paper treats robot calibration as an offline SLAM problem, where scanning poses are linked to a fixed point in space by a moving kinematic chain.
As such, the presented framework allows robot calibration using nothing but an arbitrary eye-in-hand depth sensor, thus enabling fully autonomous self-calibration without any external tools.

My new approach is utilizes a modified version of the Iterative Closest Point algorithm to run bundle adjustment on multiple 3D recordings estimating the optimal parameters of the kinematic model.
A detailed evaluation of the system is shown on a real robot with various attached 3D sensors.
The presented results show that the system reaches precision comparable to a dedicated external tracking system at a fraction of its cost.
\end{abstract}


\IEEEpeerreviewmaketitle

\section{Introduction}

\IEEEPARstart{I}{n} 2018 the American Automobile Association published a report about the repair cost of modern cars with \textit{Advanced Driver Assistance Systems} (ADAS), indicating that repairs are two to three times more expensive than for traditional cars \cite{american2018new}.
An extra charge, not only caused by having to replace the additional integrated sensors, but also by their calibration which requires dedicated equipment and specially trained personnel.
In a similar manner Richardson et al.\,\cite{richardson2013aprilcal} performed a survey on camera calibration, comparing the achieved precision reached by laymen and experts.
Their findings confirm the necessity for qualified personnel as the quality of a calibration heavily depends on capturing sufficient and evenly distributed footage of the used calibration object over the entire image area.
While there are no similar studies available for robot calibration, it is reasonable to assume that the effects with respect to cost and required know-how are similar.

\begin{figure}[bht]
	\centering%
	\includegraphics[width=\linewidth]{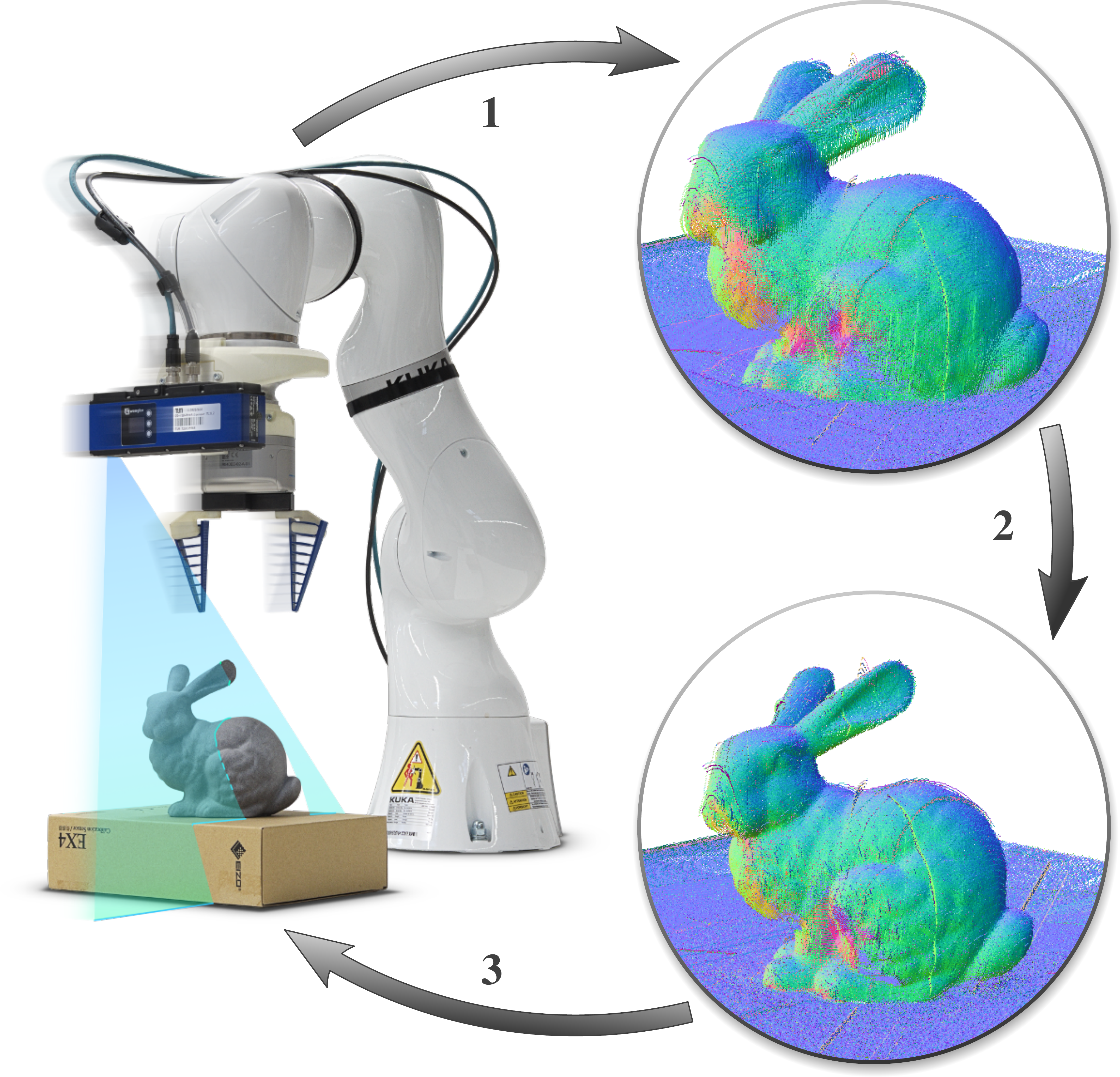}
	\caption{
		Visualization of the proposed calibration pipeline: 1) A robot with an attached 3D sensor captures multiple recordings of an arbitrary scene by moving the eye-in-hand sensor. 2) ICP based bundle adjustment is used to un-distort and align the point clouds by optimizing the robot's kinematic model. Note how the crispness improves on the bunny's ear and crib. 3) The obtained calibration parameters are uploaded to the robot manipulator.
	}%
	\label{fig:pipeline}
\end{figure}

The biggest drawback of current calibration techniques is, however, that they rely on specialized equipment.
While many approaches from literature are presented to be \enquote{autonomous} or \enquote{automated}, they can only be used when additional calibration objects, markers and/or sensors have been placed next to the robot beforehand, i.e.\,dot \cite{zhang2017computationally} or checkerboard patterns \cite{tsai1989new}, spheres \cite{li2008calibration, yin2014development, bi2017extrinsic, chen2018extrinsic}, pins \cite{wagner2015self}, or other specifically designed calibration targets \cite{antone2007fully, andersen2014calibration}.
Therefore, rendering it impossible to re-calibrate already deployed systems in unpredictable environments, such as traffic, households or catastrophic scenarios.
Though calibration often wrongly treated as a once-in-lifetime action, the system parameters are changing over a system's usage period, caused by wear-and-tear, maintenance and repairs, changes in temperature or mechanical stress i.e.\,caused by shipping or collisions.
The resulting consequences of wrong parameters may range from small imprecisions, over task failure to a potential loss of an entire system, when deployed to a hazardous environment of which it cannot escape by itself anymore.

To overcome the aforementioned issues this paper presents a framework allowing true on-site self-calibration of a robot system equipped with an arbitrary eye-in-hand 3D sensor.
Instead of using external utilities it is based on point cloud registration techniques to fuse multiple scans of a given scene.
Our approach extends the \textit{Iterative Closest Point} (ICP) algorithm to find the optimal parameters of a kinematic chain including all calibratable parameters of a robot manipulator as well as the hand-to-eye transformation (see Fig.~\ref{fig:pipeline}).
The key-contributions of this paper are:

\begin{enumerate}
	\item To the best of the author's knowledge this is the first approach solving the calibration of an entire robot system by relying only on depth data instead of external tools and objects.
	\item The presented framework allows calibration of any kinematic chain and depth sensor combination, e.g.\,single beam LiDARs, line scanners and depth cameras.
	\item A detailed evaluation is presented, comparing multiple real-world hardware configurations to a calibration performed by using traditional methods with a dedicated 3D tracking system.
\end{enumerate}
\section{Fundamentals}

By today, various technologies for measuring 3D information can be found on the market.
The most common ways to contactlessy estimate the distance between a sensor and a surface it is pointed at, are stereo vision \cite{cyganek2011introduction, sturm2011camera, ma2012invitation}, \textit{structured light} (SL) \cite{zhang2018high} and the \textit{Time-of-Flight} (ToF) principle \cite{hansard2012time}.
Depending on the used sensor it is further possible to perform multiple measurements simultaneously.
According to the used sensor model the captured range or depth measurements captured at a moment $t$ can be projected to a set of points
\begin{align}
	\phantom{\vec{p}_i}
	&\begin{aligned}
	\mathllap{P_t} &= \{ \vec{p}_1, \vec{p}_2, \cdots, \vec{p}_n \}
	\end{aligned}\\
	\intertext{with}
	&\begin{aligned}
	\mathllap{\vec{p}_i} &= \transpose{(x_i, y_i, z_i )} .
	\end{aligned}
\end{align}
Each $\vec{p}_{i}$ is a 3D point relative the sensor's origin.
As the the sensor is assumed to be mounted in an eye-in-hand configuration, its origin also forms the \textit{end effector} (EE) frame $\mathcal{E}$ of the robot's kinematic chain.
In the following work, coordinate frames of points will be denoted in superscript, calligraphic letters, such as in $\boldsymbol{p}^\mathcal{E}_i$.

When moving the robot the relative pose of $\mathcal{E}$ to the robot's base $\mathcal{B}$ changes.
Thus, one possible strategy for fusing multiple point clouds $P^\mathcal{E}_0, \cdots, P^\mathcal{E}_n$ is to transform them to $\mathcal{B}$.
The required transformation
\begin{equation}
	\transform{\mathcal{E}}{\mathcal{B}}_t =
	\begin{bmatrix}
	\mat{R} & \vec{t} \\
	\vec{0}^\textrm{T} & 1 \\
	\end{bmatrix}
\end{equation}
can be computed from the kinematic parameters $\vec{k}$ of the robot and its joint states $\vec{j}_t$ at the time $t$ at which the scan was recorded:
\begin{equation}
	\transform{\mathcal{E}}{\mathcal{B}}_t = \func{tr}(\vec{k}, \vec{j}_t),
\end{equation}
so that
\begin{equation}
	\begin{pmatrix} \vec{p}^\mathcal{B}_{i} \\ 1 \end{pmatrix} = 
	\func{tr}(\vec{k}, \vec{j}_t)
	\begin{pmatrix} \vec{p}^\mathcal{E}_{i} \\ 1 \end{pmatrix} \quad
	\forall\ \vec{p}^\mathcal{E}_{i} \in P^\mathcal{E}_{t}.
\end{equation}

While $\mat{j}_t$ is simply the sequence of the joint positions along the used robot from base to EE
\begin{equation}
	\vec{j}_t = \transpose{(j_{1,t}, j_{2,t}, \cdots, j_{n,t})},
\end{equation}
the definition of $\vec{k}$ is not as straightforward: A kinematic model suitable for calibration should be continuous and represent a complete set of \textit{Degrees of Freedom} (DoF) while staying non-redundant.
The probably still most famous way of modeling robot manipulators is by following the \textit{Denavit-Hartenberg} (DH) convention \cite{denavit1955kinematic}.
As modeling a kinematic chain with 6\,DoF per transformation includes several redundancies (e.g.\,it would be possible to shift a rotation joint along its axis) the DH convention defines joints to move along their local $z$ axis and uses only four parameters $\phi_n, d_n, a_n$ and $\alpha_n$ per segment $n$, where $\phi_n$ is the rotation around $z_{n-1}$, $d_n$ the translation along $z_{n-1}$, $a_n$ the translation along $x_n$ and $\alpha_n$ the rotation around $x_n$.

Unfortunately the DH model suffers from multiple drawbacks: It is neither complete nor parametrically continuous.
To overcome this issues Stone \cite{stone1987kinematic} suggested to use two additional parameters $b_n$ and $\gamma_n$ per segment in his \textit{S-model}, making the model complete, but not parametrically continuous at the cost of introducing redundancies.

Zhuang et Al.\,presented an alternative approach named \textit{Complete and Parametrically Continuous} (CPC) model \cite{zhuang1992complete}, which they later refined to the more intuitive \textit{Modified-CPC} (MCPC) model \cite{zhuang1993error}.
Similar to the DH convention the MCPC model assumes all joints to move along their $z$ axes.
It is constructed for each segment of the robot's kinematic chain, connecting two joints $\mathcal{J}_i$ and $\mathcal{J}_{i+1}$, by rotating the frame $\mathcal{J}_i$ around its $x$ and $y$ axes to align it's $xy$-plane with the one of $\mathcal{J}_{i+1}$ and then shift it along the new $x$ and $y$ axes to position the new origin on the $z$-axis if $\mathcal{J}_{i+1}$.

The MCPC model defines $\func{tr}(\vec{k}, \vec{j}_t)$ to be the product of alternating transformations along static segments and joints:
\begin{equation}
	\func{tr}(\vec{k}, \vec{j}_t) = \func{st}(\transpose{\vec{s}_0}) \cdot \func{jt}(j_1) \cdot \func{st}(\transpose{\vec{s}_1}) \cdot ... \cdot \func{jt}(j_n) \cdot \func{st}(\transpose{\vec{s}_n})
\end{equation}
with
\begin{equation}
\vec{k} = \transpose{(\transpose{\vec{s}}_0, \transpose{\vec{s}}_1, \transpose{\vec{s}}_2, \dots, \transpose{\vec{s}}_n)},
\end{equation}
where $\vec{s}_i$ are the parameters of the $i$-th joint.
The MCPC model uses a total of four parameters per revolute joint, two per prismatic joint and six DoF for the transformation between the last joint and the robot's EE.
For revolute joints, each segment is defined by $\vec{s} = \transpose{(\alpha, \beta, x, y)}$ so that:
\begin{equation}
	\func{st}(\transpose{\vec{s}}) = \textrm{rot}(\boldsymbol{u}_\textrm{x}, \alpha) \cdot \textrm{rot}(\boldsymbol{u}_y, \beta) \cdot \textrm{trans}(x, y, 0)
\end{equation}
where $\textrm{rot}(\boldsymbol{a}, \alpha)$ is a rotation of $\alpha$ around axis $\boldsymbol{a}$ and $\textrm{trans}(x, y, z)$ a translation along $(x,y,z)^\textrm{T}$.
$\boldsymbol{u}_x$ denotes the unit vector of a coordinate frames local $x$~axis; $\boldsymbol{u}_y$ and $\boldsymbol{u}_z$ for the $y$ and $z$~axes accordingly.
For prismatic joints the parameters $x_i$ and $y_i$ are treated as zero.

One special case it the transformation between the last joint and the EE, which has two additional degrees of freedom $\gamma$ and $z$, so that
\begin{equation}
	\small
	\func{st}(\transpose{\vec{s}_n}) = \textrm{rot}(\boldsymbol{u}_x, \alpha_n) \cdot \textrm{rot}(\boldsymbol{u}_y, \beta_n) \cdot \textrm{rot}(\boldsymbol{u}_z, \gamma) \cdot \textrm{trans}(x_n, y_n, z).
\end{equation}

\section{Related Work}

The presented approach combines techniques from two different fields of research: 1) Point cloud registration, which is commonly used in computer vision, e.g.\,for 3D reconstruction and/or \textit{simultaneous localization and mapping} (SLAM) as well as 2) solving the hand-eye and/or robot calibration problems, where especially the second one is more common the field of control engineering.
As such the state of art in both fields will be presented separately.

\subsection{Point Cloud Registration}

Regardless of its $30^\text{th}$ anniversary, the widest known approach for aligning two point clouds is still the ICP algorithm, which was developed independently by Besl and McKay \cite{besl1992method}, as well as Chen and Medioni \cite{chen1992object} in 1992.
It aims to find the rigid transformation to align one point cloud ("data") with a second one ("model"), by iteratively searching pairs of closest points between both clouds and optimizing the initially guessed transformation by minimizing the distance of all pairs.
While the general idea of both ICP versions is the same, Besl and McKay used the squared Cartesian distance of matching points as an error measure, while Chen and Medioni optimized the point-to-plane distance, which was later shown to reach a faster convergence \cite{rusinkiewicz2001efficient}.

Over the last decades numerous variants of the ICP algorithm haven been developed, using different strategies for point matching, introducing an additional validation step for point matches, and/or varying metrics as well as optimization techniques.
Detailed overviews of shape matching are given in the survey papers \cite{diez2015qualitative} and \cite{pomerleau2015review} from 2015.
An older survey of Rusinkiewicz \cite{rusinkiewicz2001efficient} from 2001 even performed a benchmark of different ICP variants.
He also suggests that \textit{Iterative Corresponding Point} might be a better fit for the ICP acronym, since many other matching criteria (e.g.\,features or backprojection) than pure geometrical distance have been shown.

Two interesting and more recent extensions of the ICP algorithm come from Segal, Haehnel and Thrun \cite{segal2009generalized}, which formulated a plane-to-plane distance function as well as from Rusinkiewicz \cite{rusinkiewicz2019symmetric} introducing a symmetric objective cost metric.
Other recent works try to extend the ICP algorithm to support non-rigid shape matching \cite{amberg2007optimal, brown2007global, cheng2017statistical} or get rid of the required initial guess \cite{attia2017efficient, cop2018delight}.
On top researchers have applied deep learning for one or multiple steps of the algorithm \cite{wang2019deep}, as well as to solve the problem of point cloud registration solely by machine learning \cite{wang2019prnet, yu2021cofinet}.

\subsection{Calibration}

Calibrating a robot system with a hand and an eye can be broken down to three separate problems: 
1) Intrinsic calibration of the optical sensor, 2) finding the transformation between the actuator and the eye and 3) calibration of the actuator itself.
A partitioning already made by Tsai and Lenz who presented a series of papers solving each step separately \cite{tsai1989overview}.

\addvspace{\subsubsectionspace}
\subsubsection{Sensor Calibration}

The issue of sensor calibration is naturally depending on the sensor and commonly treated as a standalone problem.
Even works combining approaches for solving entire system calibration at once, usually treat sensor calibration as a independent step in the overall calibration pipeline (such as \cite{fuchs2008extrinsic, miseikis2016automatic, birbach2015rapid}).

Sensor calibration must obviously fit the used device.
Especially camera calibration has become a standalone research field.
A early approach to classic camera calibration was the Eight Point algorithm \cite{longuet1981computer}, which only became stable after introducing an additional normalization procedure \cite{hartley1997defense}.
Another widely used calibration approach is the method of Zhang \cite{zhang2000flexible}.
Later works introduced more complex sensor models i.e.\,by including additional parameters for modeling radial distortion \cite{kannala2006generic, wang2008new}.

Also in the context of 3D perception more complex sensor models are required.
For stereo vision systems the camera parameters of both sensors as well as the calibration between them needs to be known, while in SL and ToF systems one needs to consider the projector instead of a second camera.
In \cite{fuchs2008extrinsic} the latency of a ToF sensor's projector is calibrated and \cite{yamazoe2012easy} demonstrates that the projector of a Kinect v1 sensor can be modeled and calibrated with in a similar fashion as a camera.
Finally \cite{muhammad2010calibration} and \cite{atanacio2011lidar} investigate the calibration of rotation multi-beam LiDAR sensors.

\addvspace{\subsubsectionspace}
\subsubsection{Eye-to-Hand Calibration}

In contrast to sensor calibration, the problems of robot calibration and eye-to-hand calibration are strongly coupled.
While there are many works assuming the actuator to be already calibrated and only focus on finding the transformation between sensor and robot (e.g.\,\cite{tsai1989new, andreff1999on-line, antone2007fully, li2008calibration, andersen2014calibration, yin2014development, heller2015globally, wagner2015self, bi2017extrinsic, zhang2017computationally, chen2018extrinsic}), calibration approaches for an entire robot often include the eye-to-hand transformation as just another robot segment.
All aforementioned calibration attempts for eye-to-hand calibration further rely on dedicated calibration objects, such as dot \cite{zhang2017computationally} or checkerboard patterns \cite{tsai1989new}, spheres \cite{li2008calibration, yin2014development, bi2017extrinsic, chen2018extrinsic}, a pin \cite{wagner2015self}, as well as specifically deigned calibration targets such as a pyramid \cite{antone2007fully} and a board with a cut-out triangle \cite{andersen2014calibration}.

More recent works tried to remove the requirement for special targets: \cite{carlson2015six} calibrated the eye-to-hand transformation by measuring generic planes, while \cite{xu2022hand-eye} used straight edges of random objects.
Heide et Al.\,\cite{heide2018calibration} estimated the pose of external LiDAR scanners by detecting a CAD based 3D model of an excavators arm in the recorded point clouds.
Sheehan et Al.\,\cite{sheehan2014automatic} managed to intrinsically self-calibrate a multi-LiDAR scanning device without an knowledge or requirements to the environment.
They defined a \textit{crispness} error metric based on squared Renyi entropy, optimizing multiple overlying scans for crisp edges.
In \cite{alismail2012automatic} and \cite{alismail2015automatic} Alismail et Al.\,solved the intrinsic calibration (4\,DoF) of a self-build 3D LiDAR made from a rotation 2D scanner, by applying ICP on data from the first and second half of a single rotation.
We recently extended this idea to show that extrinsic calibration of an eye-in-hand LiDAR is possible via fusing two 3D scans, taken at random manipulator configurations by rotating the wrist joint \cite{peters2020extrinsic}.
A similar idea was later on presented by Li et Al.\,where \textit{Particle Swarm Optimization - Gaussian Process} (PSO-GP) was used to fuse data of a close-range eye-in-hand line scanner, as ICP \enquote{is difficult to directly apply to the calibration of line laser sensors because the line laser sensors do not have the enough scanning range} \cite[Page 2]{li2021robot}---Quite in contrary to the findings of this paper: The experimental observations presented in section~\ref{sec:evaluation} actually indicate, that the calibration precision is even higher when using ICP on smaller and less complex scenes.

\addvspace{\subsubsectionspace}
\subsubsection{Robot Calibration}

While target-less calibration eye-to-hand calibration of depth sensors has been demonstrated, there are no standalone solutions for entire robot calibration yet.
General overviews to the problem of robot calibration are given in \cite{abderrahim2004accuracy} and \cite{chengang2014review}.
Examples for pure robot calibration are rather rare: Bennet and Hollerbach \cite{bennett1991autonoumous} create a loop closure in a kinematic chain, by connecting two robots of the same type at their EEs, while \cite{lightcap2008improved} and \cite{mustafa2008kinematic} use a \textit{Contact Measuring Machine} (CMM) and a precisely manufactured reference fixture to measure the position of the EE.

Another series approaches is using external, optical measuring systems.
These works usually treat the transformation between the EE and the sensor and/or markers as yet another segment of the unknown kinematic chain and include it in the calibration problem, e.g.\,by using a theodolite and a reflector mounted to the robot's EE \cite{judd1990technique} or camera/marker based 3D tracking systems \cite{jang2001calibration, ozguner2020camera}.
\cite{birbach2015rapid} calibrates the manipulator of a humanoid-like robot by watching a marker on its wrist.
Ma et Al.\,\cite{maye2015online} demonstrated a first makerless solution by using deep learning to guess both, a manipulator's kinematic model and configuration from watching it with an external camera.

Finally, there is a number of works solving robot calibration by using eye-in-hand devices.
The approaches tend to follow a similar approach as for eye-to-hand calibration.
A calibration target is recorded from a number of manipulator poses.
Using the constraints that neither the target, nor the robot's base have moved it is possible to to calculate the optimal parameters for the kinematic model.
Even the used targets are often the same: \cite{strobl2006optimal} and \cite{pradeep2014calibrating} use a checkerboard while \cite{knoll2000einrichtung, yu2018simultaneous} and \cite{wang2020point} use one or multiple spheres.
\cite{kang2007autonomous} calibrated a robot with an attached line scanner by following a spanned string and in \cite{ruther2010narcissistic} the robot system detects markers at its joints by watching itself in a mirror.
\section{Approach}

The aim of this work is to find the optimal parameters $\boldsymbol{k}_\textrm{opt}$ describing the kinematic model of a robot.
In a first step multiple scans of the robot's environment are recorded by moving a depth sensor attached to its EE.
By exploiting the knowledge about the robot's design and configuration at the time of scanning, one can transform all acquired point cloud data to $\mathcal{B}$.
In a static environment all overlapping points from the performed scans must match to the same surfaces.
Thus any errors in the fused reconstructions must originate from errors in the projection of measurements along the kinematic chain.
By formulating a cost function to express the quality of matching points, we can use the ICP algorithm to minimize the projection error and thus estimate $\boldsymbol{k}_\textrm{opt}$.

\subsection{Data Structure and Notation}

Lets consider a robot with an actuated depth sensor, i.e.\,in an eye-in-hand configuration.
As the used measuring technique and lens model may differ, for the scope of this work the sensor is assumed to be intrinsically calibrated and to produce a set of points $P$.
While it is negligible how the data was measured, it is essential to known when each depth value was obtained.
Depending on the sensor type one obtains a different amount of data at a time $t$.
In detail, a single beam LiDAR only measures a single range value, a triangulation based line scanner captures a vector of points and a depth camera even returns an entire matrix of measurements.
Thus, depending on the used sensor type the elements of $P$ can be rearranged to form a matrix $\mat{P}$:
\begin{align}
	\phantom{\mat{P}^\textrm{Depth Camrea}}
	&\begin{aligned}
		\mathllap{\mat{P}^\textrm{LiDAR}} &= \vec{p} ,
	\end{aligned}\\
	&\begin{aligned}
		\mathllap{\mat{P}^\textrm{Line Scanner}} &= ( \vec{p}_1, \vec{p}_2, ... , \vec{p}_n )^\textrm{T},
	\end{aligned}\\
	\intertext{and}
	&\begin{aligned}
		\mathllap{\mat{P}^\textrm{Depth Camera}} &=
		\begin{bmatrix}
			\vec{p}_{1,1} & \vec{p}_{1,2} & \cdots & \vec{p}_{1,m}\\
			\vec{p}_{2,1} & \vec{p}_{2,2} & \cdots & \vec{p}_{2,m}\\
			\vdots               & \vdots               & \ddots & \vdots \\
			\vec{p}_{n,1} & \vec{p}_{n,2} & \cdots & \vec{p}_{n,m}\\
		\end{bmatrix} .
	\end{aligned}%
\end{align}
Recording multiple scans while moving the robot finally combines multiple $\mat{P}$ to a dataset tensor $\mat{D}$:
\begin{align}
	\phantom{\mat{D}^\textrm{Depth Camrea}}
	&\begin{aligned}
	\mathllap{\mat{D}^\textrm{Line Scanner}} &= 
	\begin{pmatrix}
		\mat{P}^{\textrm{Line Scanner}^T}_1\\
		\mat{P}^{\textrm{Line Scanner}^T}_2\\
		\vdots\\
		\mat{P}^{\textrm{Line Scanner}^T}_m
	\end{pmatrix}^\textrm{T}
	\end{aligned}\\
	\intertext{and}
	&\begin{aligned}
	\mathllap{\mat{D}^\textrm{Depth Camrea}} &= 
	\begin{pmatrix}
	\mat{P}^\textrm{Depth Camrea}_1\\
	\mat{P}^\textrm{Depth Camrea}_2\\
	\vdots\\
	\mat{P}^\textrm{Depth Camrea}_m
	\end{pmatrix}.
	\end{aligned}
\end{align}
Rotating LiDARs provide one special case, as they also measure a line but sequentially.
To maintain the spatial order of points in a LiDAR scan, all $l$ points from a single rotation are arranged column wise, so that
\begin{equation}
	\mathllap{\mat{D}^\textrm{LiDAR}} = 
	\begin{bmatrix}
	\mat{P}^\textrm{LiDAR}_{1,1} & \mat{P}^\textrm{LiDAR}_{1,2} & \cdots & \mat{P}^\textrm{LiDAR}_{1,m}\\
	\mat{P}^\textrm{LiDAR}_{2,1} & \mat{P}^\textrm{LiDAR}_{2,2} & \cdots & \mat{P}^\textrm{LiDAR}_{2,m}\\
	\vdots & \vdots & \ddots & \vdots\\
	\mat{P}^\textrm{LiDAR}_{l,1} & \mat{P}^\textrm{LiDAR}_{l,2} & \cdots & \mat{P}^\textrm{LiDAR}_{l,m}\\
	\end{bmatrix}.
\end{equation}

For all measurements $\mat{P}_t$ there are matching vector with the robot's joint states $\boldsymbol{j}_t$.
The issue of different measuring frequencies can be overcome by interpolation of the joint positions to find an approximation of the exact state for $t$.
As a result there is a matching a matrix
\begin{equation}
	\mat{J}_i = (\boldsymbol{j}_1, \boldsymbol{j}_2, \cdots, \boldsymbol{j}_m)
\end{equation}
matching each $\mat{D}_i$, or a tensor
\begin{equation}
	\mathllap{\mat{J}^\textrm{LiDAR}_i} = 
	\begin{bmatrix}
		\vec{j}_{1,1} & \mat{j}_{1,2} & \cdots & \mat{j}_{1,m}\\
		\vec{j}_{2,1} & \mat{j}_{2,2} & \cdots & \mat{j}_{2,m}\\
		\vdots & \vdots & \ddots & \vdots\\
		\mat{j}_{l,1} & \mat{j}_{l,2} & \cdots & \mat{j}_{l,m}\\
	\end{bmatrix}.
\end{equation}
for each $\mat{D}^\text{LiDAR}_i$, respectively.

For the following formulations the recorded data will be denoted as $\mat{D}$, regardless of the used sensor type.
While the tensor $\mat{D}^\textrm{Depth Camrea}$ has a rank of three (four if one counts each point's $x,y,z$ coordinates), I use the two-index notation $\vec{p}_{i,j}$ as an equivalent to
\begin{equation}
\vec{p}_{i,j} = \vec{p}_{i,j,k} \forall k
\end{equation}
for simplification.
In the same way I use $\boldsymbol{j}_{i,j}$ to address the joint states matching a point $\vec{p}_{i,j}$, even though $\boldsymbol{J}$ may have a different rank than $\mat{D}$.

\subsection{Parameter Modeling}

I use the MCPC model to express the spatial relationships of the robots kinematic chain.
An initial guess $\boldsymbol{k}_\textrm{init}$ of $\boldsymbol{k}_\text{opt}$ can i.e.\,be obtained via manual measuring or from the robot's datasheet.
Though the model MCPC itself is free of redundancies there are still a few parameters which cannot be calibrated.
The transformation between $\mathcal{B}$ and the first joint of th robot $\mathcal{J}_1$ behaves as a static offset to all recorded point clouds and there is no information to deduce its parameters.
Since the frame of the first joint $\mathcal{J}_1$ is lacking a fixed position in space, there further arise two redundancies in the parameters of $\mat{T}^{\mathcal{J}_2 \rightarrow \mathcal{J}_1}$.
As such the $y$ and $\beta$ also have to be excluded from the optimization.
I thus use a bitmask vector $\vec{m}$ of the same size as $\boldsymbol{k}$ with 
\begin{equation}
	\vec{m} = (m_1, m_2, \cdots, m_n) \mid m_i \in \{0,1\} \forall i
\end{equation}
to define which parameters shall be taken into account in the optimization process.
$\vec{m}$ is also used to reduce the number of DoFs for prismatic joints.

\subsection{Normalization}

The influence of orientation errors on the cost metric heavily depends on the distance of surface to the sensor.
To balance the weight of angular and translation parameters $\boldsymbol{k}$ should be normalized before optimization.
Since all data is captured by the robot looking at it's environment, one can assume the data to be more or less equally distributed around the robot's base.
Thus only approximating the scaling factor $s$ from a subset of all datasets is usually sufficient.
I project all points of the first recorded dataset $\mat{D}^\mathcal{E}_1$ to $\mathcal{B}$ to obtain a tensor of points $\mat{D}^\mathcal{B}_1$ with
\begin{equation}
	\label{eq:projection}
	\vec{p}^\mathcal{B}_{i,j} = \func{tr}(\vec{k}_\idx{init}, \vec{j}_{i,j}) \vec{p}^\mathcal{E}_{i,j} \quad
	\forall \vec{p}^\mathcal{B}_{i,j} \in \mat{D}^\mathcal{B}_1, \vec{p}^\mathcal{E}_{i,j} \in \mat{D}^\mathcal{E}_1,
\end{equation}
allowing to compute $s$ by
\begin{equation}
	s = \frac{1}{n} \sum_{i=1}^{n} {\| \vec{p}^\mathcal{B}_i \|} \quad
	\mid \vec{p}^\mathcal{B}_i \in \mat{D}^\mathcal{B}_1.
\end{equation}

I denote components normalized by $s$ with a hat symbol (as in $\hat{\vec{p}}$).
The normalization of a dataset $\mat{D}^\mathcal{E}$ is straightforward:
\begin{equation}
	\boldsymbol{\hat{D}}^\mathcal{E} = \frac{1}{s} \mat{D}^\mathcal{E}.
\end{equation}
Note that one also needs to normalize the translation parameters of $\boldsymbol{k}$
\begin{equation}
	\small
	\boldsymbol{\hat{k}} = (\alpha_0, \beta_0, \frac{x_0}{s}, \frac{y_0}{s}, \alpha_1, \beta_1, \frac{x_1}{s}, \frac{y_1}{s}, \cdots, \alpha_n, \beta_n, \gamma, \frac{x_n}{s}, \frac{y_n}{s}, \frac{z}{s})^\textrm{T}
\end{equation}
as well as the joint positions of prismatic joints.

Even though the parameters of $\boldsymbol{k}$ change with every ICP iteration, it is safe to keep $s$ constant over the whole optimization process.
As the accumulated error of $\boldsymbol{k}_\textrm{init}$ is usually in the range of a few centimeters the numerical effects on $s$ are neglectable.

In case the initial assumption of an equal distribution of points around the robot does not hold, it might be necessary to a) compute $s$ by averaging points from all datasets, as well as b) to add an additional translation $\transform{\mathcal{B}}{\mathcal{C}}$ shifting the points to their average centerpoint, which would need to be incorporated as an additional transformation in $\func{tr}(\vec{k}, \vec{j})$ and masked in $\vec{m}$.

\subsection{ICP}

\begin{figure*}[bht]
	\centering%
	\includegraphics[width=0.33\linewidth]{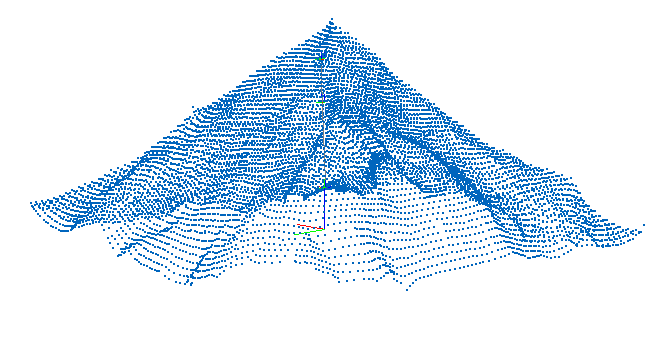}%
	\includegraphics[width=0.33\linewidth]{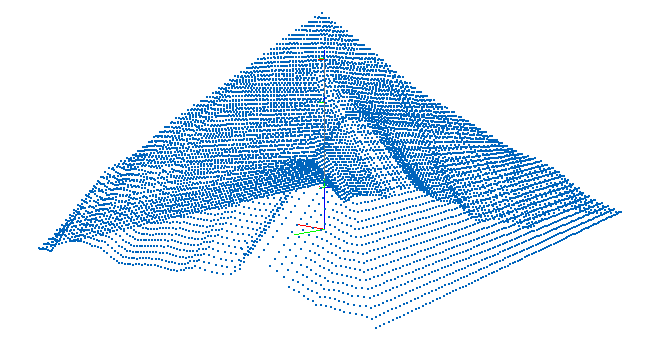}%
	\includegraphics[width=0.33\linewidth]{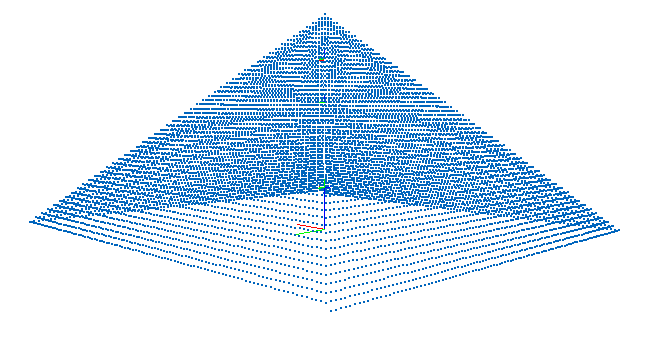}%
	\caption{
		Distortion of point clouds obtained by projecting depth measurements of a pyramid along an imprecise kinematic model with randomized scanning trajectories.
		From left to right: One single point per joint configuration (single-beam LiDAR), one line per configuration (line scanner) and a matrix of points obtained from a single configuration (depth camera).
	}%
	\label{fig:distortions}
\end{figure*}

The used ICP algorithm uses four steps which I will describe in detail below: 1) Initial projection and validation of points, 2) search for point matches, 3) verification of point matches and 4) computation of the error function and optimization of $\boldsymbol{\hat{k}}$.
All four steps are repeated iteratively until convergence is reached.
A detailed overview of the used operations is given in algorithm~\ref{alg_icp}.

\begin{algorithm}[htb]
	\caption{ICP for Calibration}
	\begin{algorithmic}
		\Require Datasets $\set{\mat{\hat{D}}^{\mathcal{E}}_1, \mat{\hat{D}}^{\mathcal{E}}_2, \cdots, \mat{\hat{D}}^{\mathcal{E}}_n}$
		\Require Joint positions $\set{\mat{\hat{J}}^{\mathcal{E}}_1, \mat{\hat{J}}^{\mathcal{E}}_2, \cdots, \mat{\hat{J}}^{\mathcal{E}}_n}$
		\Require Initial guess $\mat{k}_\idx{init}$ and parameter mask $\vec{m}$
		\Require Stop threshold $\epsilon$
		
		\State $\vec{\hat{k}}_\idx{opt} \leftarrow \func{normalizeTranslation}(\vec{k}_\idx{init})$
		\Repeat
			\State Compute $\mat{\hat{D}}^{\mathcal{B}}_1, \mat{\hat{D}}^{\mathcal{B}}_2, \cdots, \mat{\hat{D}}^{\mathcal{B}}_n$:\, $\vec{\hat{p}}_t^\mathcal{B} = \func{tr}(\vec{\hat{k}}_\idx{opt}, \vec{\hat{j}}_t) \vec{\hat{p}}_t^\mathcal{E} \; \forall t$
			\State $\mat{\hat{O}}^\mathcal{B}_i \leftarrow \func{detectNoiseAndEdges}(\mat{\hat{D}}^{\mathcal{B}}_i) \; \forall i$
			\State $\mat{\hat{Q}}^\mathcal{B}_i \leftarrow \mat{\hat{D}}^\mathcal{B}_i \setminus \mat{\hat{O}}^\mathcal{B}_i \; \forall i$
			\State $M \leftarrow \{\}$
			\ForEach {$\mat{\hat{Q}}^\mathcal{B}_i, \mat{\hat{Q}}^\mathcal{B}_j \mid i < j$}
				\State $M \cup \func{findMatches}(\mat{\hat{Q}}^\mathcal{B}_i, \mat{\hat{Q}}^\mathcal{B}_j)$
			\EndFor
			\State $M \leftarrow \func{filter}(M)$
			\State Solve $\min_{\vec{\hat{k}} \mid \hat{k}_i \in \vec{\hat{k}}_\idx{opt} \wedge m_i = 1 \mid m_i \in \vec{m}} \ \func{e}(M, \vec{\hat{k}}_\idx{opt})$
			\State $\vec{k}_\idx{opt} \leftarrow \func{denormalizeTranslation}(\vec{\hat{k}}_\idx{opt})$
		\Until{$\abs{\Delta \vec{k}_\idx{opt}} \leq \epsilon$}
		\State \Return $\vec{k}_\idx{opt}$
	\end{algorithmic}
	\label{alg_icp}
\end{algorithm}

Note that the projection of $\mat{\hat{D}}^\mathcal{E}$ to $\mat{\hat{D}}^\mathcal{B}$ is the result of a transformation along a kinematic chain, parameterized by joint states that change over the duration of the recording.
In other words: Even though the individual transformations based on $\boldsymbol{\hat{k}}$ are rigid for themselves, errors in the kinematic parameters will lead to non-linear distortions as shown in fig.\,\ref{fig:distortions}.
An effect making it hard to apply common tools often used in point cloud registration, as features, tree-structures for point matching and normals cannot be pre-computed.
This is an especial limiting factor for the cost function, as numeric optimization of $\vec{\hat{k}}$ leads to an ongoing deformation of the resulting point cloud.
I thus use the cross product for normal computation
\begin{equation}
	\vec{n}_{i,j} = \func{n}(\vec{p}^\mathcal{B}_{i,j}) =
	\frac{\func{m}(\vec{p}^\mathcal{B}_{i-1,j}, \vec{p}^\mathcal{B}_{i,j-1}) + \func{m}(\vec{p}^\mathcal{B}_{i+1,j}, \vec{p}^\mathcal{B}_{i,j+1})}
	{\abs{\func{m}(\vec{p}^\mathcal{B}_{i-1,j}, \vec{p}^\mathcal{B}_{i,j-1}) + \func{m}(\vec{p}^\mathcal{B}_{i+1,j}, \vec{p}^\mathcal{B}_{i,j+1})}}
\end{equation}
with
\begin{equation}
	\func{m}(\vec{p}, \vec{q}) = \frac{\vec{p} \times \vec{q}}{\abs{\vec{p} \times \vec{q}}}
\end{equation}
and the point-to-plane error metric as a compromise between runtime and precision.
Since $\vec{n}_{i,j}$ is normalized, it is the same whether it is computed via $\func{n}(\vec{p}_{i,j})$ or $\func{n}(\vec{\hat{p}}_{i,j})$.%

Note that for scans other than depth images, the normal orientation depends on the scanning trajectory and the resulting order of scan lines (e.g.\,whether the lines were recorded from left to right or from right to left).
It may thus be necessary to verify the normal orientation by comparing it to the direction of the sensor's view ray:
\begin{equation}
	\vec{v}_{i,j} = \func{vn}(\vec{p}^\mathcal{B}_{i,j}) = \left\{
	\begin{array}{rl}
		\func{n}(\vec{p}^\mathcal{B}_{i,j}) & \func{n}(\vec{p}^\mathcal{B}_{i,j}) \cdot \vec{o}^\mathcal{B}_t \leq 0 \\
		-\func{n}(\vec{p}^\mathcal{B}_{i,j}) & \, \text{else} \\
	\end{array}
	\right.
\end{equation}
where the origin of the sensor $\vec{o}^\mathcal{B}_t$ at a time $t$ is simply the translational part of $\transform{\mathcal{E}}{\mathcal{B}}_t$.

\addvspace{\subsubsectionspace}
\subsubsection{Projection and Validation}

\begin{figure}[htb]
	\centering%
	\includegraphics[width=0.75\linewidth]{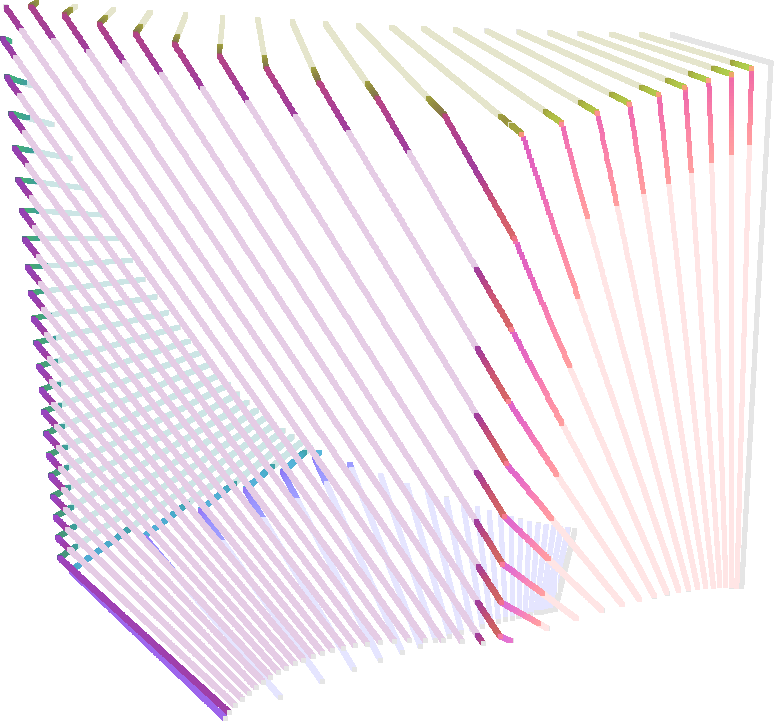}%
	\caption{
		Scan lines of a cubic room, colorized by the direction of the points' normal vectors.
		For points along corners some of the neighbouring points lie on other surfaces, causing the normals to bend toward edges.
		The effect becomes visible as a color gradient on the scan lines.
		Areas with uniform normals are rendered in lighter colors.
	}%
	\label{fig:incorrect_normals}
\end{figure}

Before searching for point matches, all datasets $\mat{\hat{D}}^{\mathcal{E}}_i$ have to be projected to $\mathcal{B}$:
\begin{equation}
	\vec{\hat{p}}^\mathcal{B}_{t} = \func{tr}(\vec{\hat{k}}_\idx{init}, \vec{\hat{j}}_{t}) \vec{\hat{p}}^\mathcal{E}_{t} \quad
	\forall \vec{\hat{p}}^\mathcal{B}_{t} \in \mat{\hat{D}}^\mathcal{B}_i, \vec{\hat{p}}^\mathcal{E}_{t} \in \mat{\hat{D}}^\mathcal{E}_i \quad \forall i .
\end{equation}

It is natural for sensor data to contain noise and outliers, as well as in the context of depth sensors gaps.
Moreover, the chosen approach for normal estimation is prone to make incorrect assumptions along corners and edges.
An affect becoming especially eminent when the density between separate scan lines and points within a single scan line heavily differs (see fig.~\ref{fig:incorrect_normals}).
To improve the quality of the calibration result it is recommendable to remove such measurements from the recorded data.

While the detection of invalid points (usually indicated by \texttt{NaN} values) is straightforward, the classification of outliers and edges requires an additional filtering routine.
A window of $n \times m$ radius is applied to each projected point $\vec{\hat{p}}^\mathcal{B}_{i,j}$ to compute overlap of neighboring normals
\begin{equation}
	\func{o}(\vec{\hat{p}}^\mathcal{B}_{i,j}) = 
	\sum_{a = -n}^{n} \sum_{b = -m}^{m} \frac{\abs{\vec{v}_{i,j} \cdot \vec{v}_{i+a,j+b}}}{(2n+1)(2m+1)}
\end{equation}
and exclude all points with a value $\func{o}(\vec{\hat{p}}_{i,j})$ below a threshold $g_\idx{min}$.
By choosing different values for $n$ and $m$ it is possible to compensate for varying densities of points and scan lines.
As the indices of points are the same in $\mat{\hat{D}}^\mathcal{B}$ and $\mat{\hat{D}}^\mathcal{E}$ one can also remove the according measurements from the raw data.
The filtered point clouds are denoted $\mat{\hat{Q}}^\mathcal{B}$ and $\mat{\hat{Q}}^\mathcal{E}$, respectively.

\addvspace{\subsubsectionspace}
\subsubsection{Point Matching}

For each pair of filtered datasets $\langle \mat{\hat{Q}}^\mathcal{B}_i, \mat{\hat{Q}}^\mathcal{B}_j \rangle$ a $k$-d tree \cite{bentley1975multidimensional} is constructed over $\mat{\hat{Q}}^\mathcal{B}_j$.
It is used to find the closest neighbor $\vec{\hat{p}}^\mathcal{B}_v \in \mat{\hat{Q}}^\mathcal{B}_j$ for every point $\vec{\hat{p}}^\mathcal{B}_u \in \mat{\hat{Q}}^\mathcal{B}_i$.
Based on the indices $u$ and $v$ a pair $\langle \vec{\hat{p}}^\mathcal{E}_i, \vec{\hat{p}}^\mathcal{E}_j \rangle$ is obtained and added to a $M$.

Since---other than the original ICP---bundle adjustment is used to align more than two scans at once, the process is repeated for all possible combinations of datasets (without respect to the order).

\begin{table*}[htb]
	\caption{Sensor characteristics and used parameters}%
	\label{tab:sensors}%
	\setlength{\tabcolsep}{0.75em}%
	\centering%
	\begin{tabular}{l rcl lr rr | ccc cc cc}
		\toprule
		& \multicolumn{3}{c}{Resolution} & \multicolumn{2}{c}{View Range} & \multicolumn{2}{c|}{Noise Ratio} & \multicolumn{3}{c}{Preprocessing} & \multicolumn{2}{c}{Match Validation} & \multicolumn{2}{c}{Stop Criteria}\\
		& $x$ & \hspace{-1.0em}$\times$\hspace{-1.0em} & $y$ & \,Closest & \multicolumn{1}{c}{Farthest} & \multicolumn{1}{c}{$\sigma_\idx{abs}$} & \multicolumn{1}{c|}{$\sigma_\idx{rel}$} & $n$ & $m$ & $g_\idx{min}$ & $d_\idx{max}$ & $f_\idx{min}$ & $\epsilon$ & $i_\idx{max}$ \\
		\midrule
		Hokuyo UTM 30LX        & $1080$ & \hspace{-1.0em}$\times$\hspace{-1.0em} & $1^\text{(a)}$   & 10.0\unit{cm} & 4000.0\unit{cm} & 18.00\unit{mm} & -- & 2 & 4 & 0.60 & 20\unit{mm} & 0.75 & $10^{-4}$ & 50 \\
		Wenglor MLSL236        & $1280$ & \hspace{-1.0em}$\times$\hspace{-1.0em} & $1$            & 30.0\unit{cm} &  150.0\unit{cm} &  0.20\unit{mm} & -- & 3 & 2 & 0.80 & 20\unit{mm} & 0.80 & $10^{-4}$ & 50 \\
		Microsoft Kinect Azure &  $320$ & \hspace{-1.0em}$\times$\hspace{-1.0em} & $288^\text{(b)}$ & 50.0\unit{cm}$^\text{(d)}$ &  546.0\unit{cm} &  2.53\unit{mm} & 0.21\% & 2 & 2 & 0.75 & 20\unit{mm} & 0.80 & $10^{-4}$ & 50 \\
		PhotoNeo MotionCam 3D  & $1120$ & \hspace{-1.0em}$\times$\hspace{-1.0em} & $800^\text{(d)}$ & 49.7\unit{cm} &   93.9\unit{cm} &  0.18\unit{mm} & -- & 2 & 2 & 0.80 & 20\unit{mm} & 0.80 & $10^{-4}$ & 50 \\
		\bottomrule
	\end{tabular}
	\vspace{0.4em}\\
	a)\hspace{0.4em}All 1080 range values are recorded sequentially during rotation%
	\hspace{2em}b)\hspace{0.4em}NFOV $2\times2$ binned mode%
	\hspace{2em}c)\hspace{0.4em}Dynamic mode\\%
	d)\hspace{0.4em}According to datasheet. However, experiments have shown that the used objects were already measurable at less than 20\unit{cm} distance.%
\end{table*}

\begin{figure*}[htb]
	\centering%
	\includegraphics[width=0.24\linewidth]{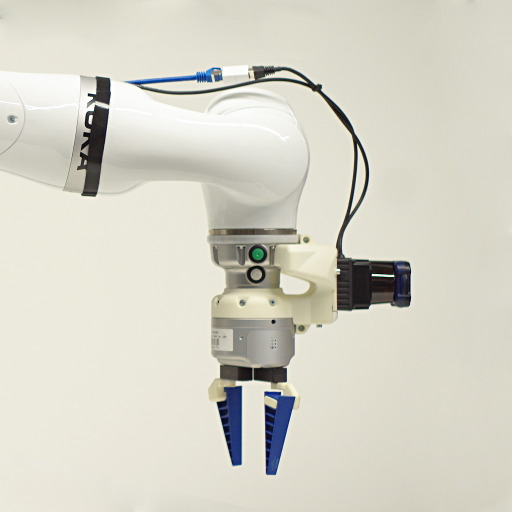}\hfill%
	\includegraphics[width=0.24\linewidth]{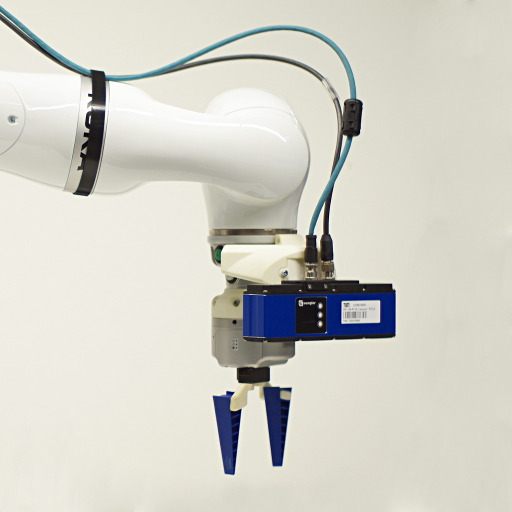}\hfill%
	\includegraphics[width=0.24\linewidth]{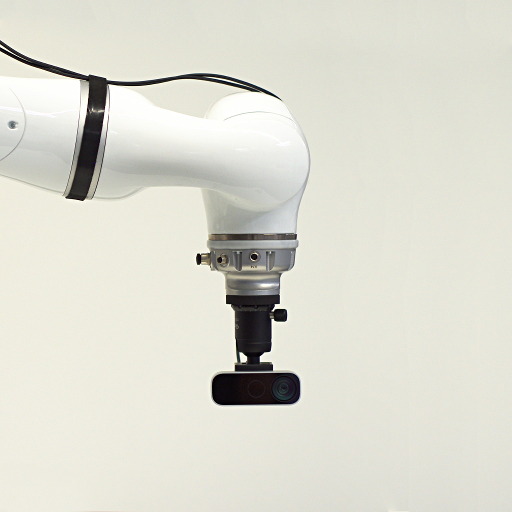}\hfill%
	\includegraphics[width=0.24\linewidth]{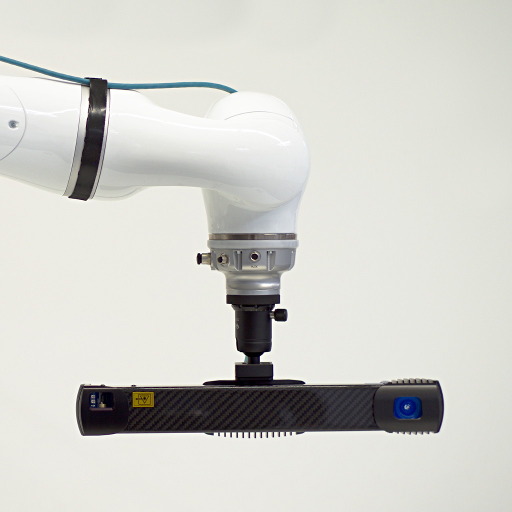}%
	\caption{
		Used sensors from left to right: 1) Hokuyo UTM-30LX laser range finder, 2) Wenglor MLSL236 line scanner, 3) Microsoft Kinect Azure and 4) PhotoNeo MotionCam 3D.
	}%
	\label{fig:sensors}
\end{figure*}

\addvspace{\subsubsectionspace}
\subsubsection{Match Validation}

Unfortunately pure nearest neighbor matching is prone to finding incorrect pairs, e.g.\,when a point $\vec{\hat{q}}^\mathcal{B} \in \mat{\mat{\hat{Q}}}^\mathcal{B}_i$ lies in on a surface not captured in $\mat{\hat{Q}}^\mathcal{B}_j$.
Thus all matches with a point-to-point distance
\begin{equation}
	\func{d}(\vec{\hat{q}}^\mathcal{B}_i, \vec{\hat{q}}^\mathcal{B}_j) =
	\abs{\vec{\hat{q}}^\mathcal{B}_i - \vec{\hat{q}}^\mathcal{B}_j}
\end{equation}
above a maximum distance $\hat{d}_\idx{max}$ and a normal overlap
\begin{equation}
	\func{f}(\vec{\hat{q}}^\mathcal{B}_i, \vec{\hat{q}}^\mathcal{B}_j) = \func{vn}(\vec{\hat{q}}^\mathcal{B}_i) \cdot \func{vn}(\vec{\hat{q}}^\mathcal{B}_j)
\end{equation}
below a threshold $f_\idx{min}$ are excluded from $M$ (similar to \cite{newcombe2011kinectfusion}).

\addvspace{\subsubsectionspace}
\subsubsection{Cost Function}

The point-to-plane distance of a match $\langle \vec{\hat{q}}^\mathcal{B}_i, \vec{\hat{q}}^\mathcal{B}_j \rangle \in M$ is defined by
\begin{equation}
	\func{c}(\mat{\hat{q}}^\mathcal{B}_i, \mat{q}^\mathcal{B}_j, \vec{\hat{k}}) =
	[ \func{tr}(\vec{\hat{k}}, \vec{\hat{j}}_i)\vec{\hat{p}}^\mathcal{E}_i -
	\func{tr}(\vec{\hat{k}}, \vec{\hat{j}}_j)\vec{\hat{p}}^\mathcal{E}_j ]
	\cdot \func{n}(\mat{\hat{q}}^\mathcal{B}_i)
\end{equation}
resulting in a total error
\begin{equation}
	\func{e}(\vec{\hat{k}}, M) = 
	\sum_{\langle \vec{\hat{q}}^\mathcal{B}_i, \vec{\hat{q}}^\mathcal{B}_j \rangle \in M} \func{c}(\mat{\hat{q}}^\mathcal{B}_i, \mat{\hat{q}}^\mathcal{B}_j, \vec{\hat{k}})^2.
\end{equation}
Since the normals are only used as part of a squared dot product, one can safely ignore its sign in the computation of the error.
The Levenberg-Marquard method \cite{marquardt1963algorithm, levenberg1944method} is used to find the optimal, unmasked parameters of $\vec{\hat{k}}_\idx{opt}$ minimizing
\begin{equation}
	\min_{\vec{\hat{k}} \mid \hat{k}_i \in \vec{\hat{k}}_\idx{opt} \wedge m_i = 1 \mid m_i \in \vec{m}} \ \func{e}(M, \vec{\hat{k}}_\idx{opt})
\end{equation}

The final calibration parameters $\vec{k}_\idx{opt}$ can then be obtained by removing the scaling factor $s$ from $\vec{\hat{k}}_\idx{opt}$:
\begin{align}
	\begin{split}
		\boldsymbol{k} = ( & \alpha_0, \beta_0, s\hat{x}_0, s\hat{y}_0,\\
		                   & \alpha_1, \beta_1, s\hat{x}_1, s\hat{y}_1,\\
		                   & \cdots,\\
		                   & \alpha_n, \beta_n, \gamma, s\hat{x}_n, s\hat{y}_n, s\hat{z})^\textrm{T}.
	\end{split}
\end{align}

All four steps of the ICP algorithm described above are iteratively repeated until $\Delta \vec{k}_\idx{opt}$ between two iterations reaches below a threshold $\epsilon$.
\section{Evaluation}\label{sec:evaluation}

\begin{figure}[htb]
	\centering%
	\def\svgwidth{0.95\linewidth}%
	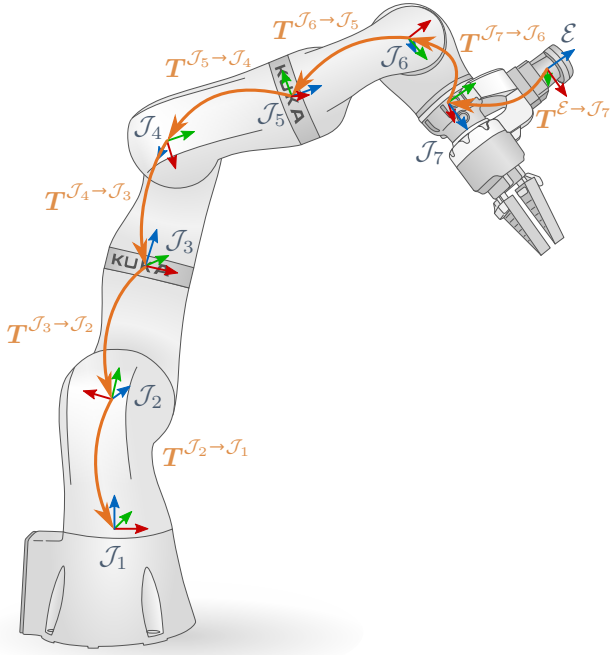%
	\caption{
		Physical joint frames of the used LBR iiwa R840 robot.
		All axes rotate around their local $z$-axes.
		The base frame $\mathcal{B}$ for the kinematic chain is equivalent to the frame of the first joint $\mathcal{J}_1$.
		When applying the MCPC modeling convention the joints $\mathcal{J}_7$, $\mathcal{J}_5$ and$\mathcal{J}_3$ are shifted along their $z$-axes to the positions of their predecessors.
		Setting the non-calibratable parameters $y$ and $\beta$ of $\mat{T}^{\mathcal{J}_1 \rightarrow \mathcal{J}_2}$ to zero also shifts $\mathcal{J}_1$ (and thus also $\mathcal{B}$) up to have $\mathcal{J}_2$ lying in its $xy$-plane.
	}
	\label{fig:iiwa_kinematics}
\end{figure}

The aforementioned formulations yield in a generic framework, theoretically enabling the calibration of any kinematic chain with an attached 3D sensor, as long as it is possible to get precise readings of the included joint positions.
To demonstrate the capabilities of the proposed system it is tested on a seven DoF robot arm in combination with varying 3D sensors.
In detail, I will use a KUKA LBR iiwa R840 manipulator (see figure~\ref{fig:iiwa_kinematics}) and a total of four sensors with different characteristics:
A Hokuyo UTM-30LX rotating single-beam LiDAR, a Wenglor MLSL236 triangulation based line scanner, a Microsoft Kinect Azure consumer grade depth camera and a PhotoNeo MotionCam 3D high end industrial grade depth camera (see figure~\ref{fig:sensors}).

The used LiDAR has by far the widest view range of all tested sensors, but is only meant for far range scanning.
Moreover the studies shown in \cite{pomerleau2012noise} suggest that the sensor error may be modeled as absolute Gaussian noise with a standard deviation of 1.8\unit{cm}.

In contrast the Wenglor MLSL236 is extremely precise, but in its view range limited to close range scenes.
Unfortunately there are no studies about its sensor model available.
However, the vendor's datasheet specifies the the maximum depth error to stay below 600\unit{$\mu$m} \cite{wenglormlsl236}.
Given the three-sigma-rule one thus may assume a Gaussian noise model with $\sigma_\idx{abs} = 0.2\unit{mm}$.

\begin{figure}[h!]
	\centering%
	\input{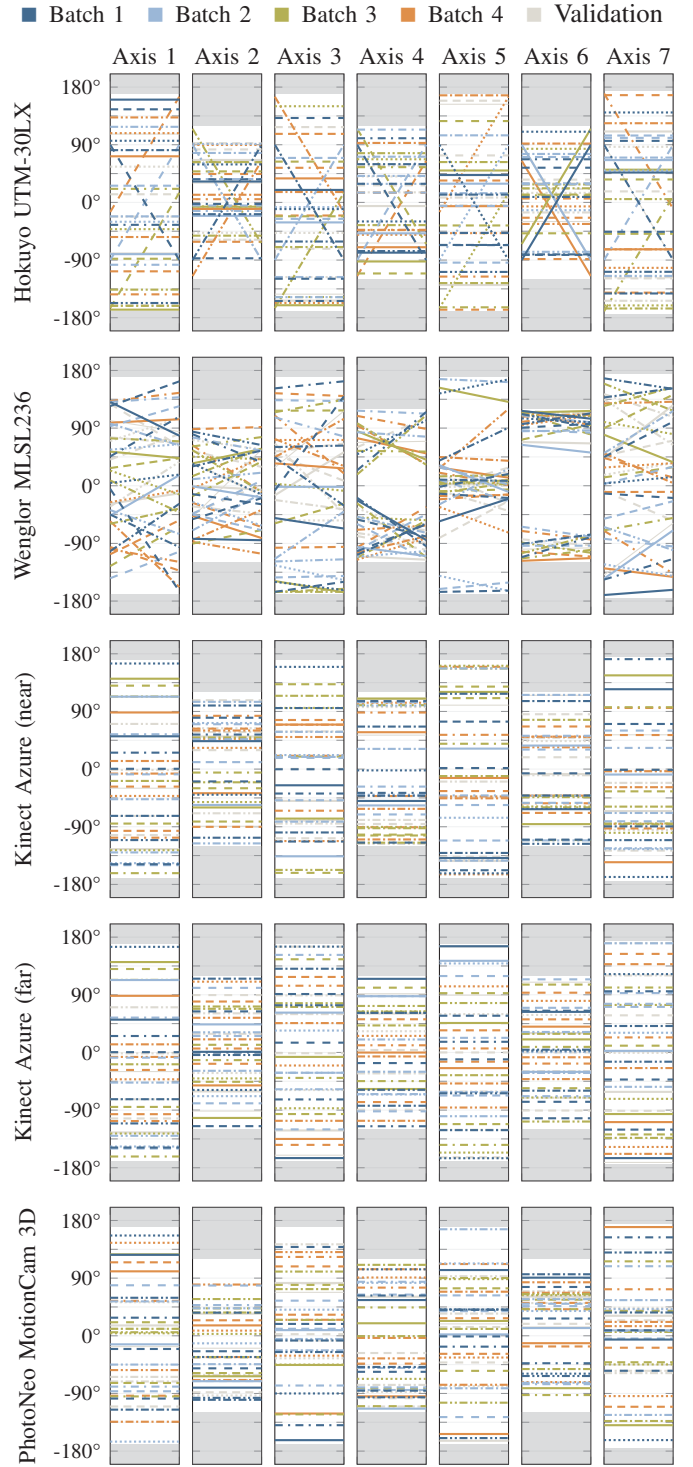}%
	\caption{
		Scanning poses and trajectories used per sensor.
		Colors are defined by batch, while the line style allows to identify a single configuration.
		Areas outside of the joint axes motion ranges are grayed out.
		The scans of the first batch were used for the test runs with seven datasets.
		Datasets from the other batches were added subsequently to increase the overall number of datasets.
	}%
	\label{fig:joint_positions}
\end{figure}

Microsoft's Kinect Azure is the only consumer grade 3D sensor in the test field and with a recommended retail price of around 400$\unit{EUR}$ by far the cheapest one tested.
It is also the only sensor suitable for both, close and far range scenes.
The Kinect offers a \textit{narrow field of view} (NFOV) as well as a \textit{wide field of view} (WFOV) scanning mode, both which can be combined with an additional $2\times2$ binning, which increases the $z$-precision at a cost of the $xy$-resolution.
All datasets of the Kinect Azure were captured in NFOV $2\times2$ binned mode.
According to the findings of \cite{tolgyessy2021evaluation} the sensor's noise model for this mode is similar to linear Gaussian noise modeled by
\begin{equation}
\sigma = \sigma_\idx{rel} z + \sigma_\idx{abs},
\end{equation}
with $\sigma_\idx{rel} = 0.21\unit{\%}$, $\sigma_\idx{abs} = 2.53\unit{mm}$ and $z$ to be the distance of a particular measurement.

Finally the PhotoNeo MotionCam 3D is a high end industrial grade depth camera for scenes between 50\unit{cm} and 1\unit{m} distance.
It can be seen as a successor to PhotoNeo's PhoXi 3D Scanner, adding support for dynamic scanning at up to 20\unit{Hz}.
As the used sensor technology is very similar to the PhoXi 3D Scanner, one may take the observations presented in \cite{cop2021new} as a point of reference for modeling the sensor noise: Averaged over four different materials $\sigma_\idx{abs}$ has a value of 0.18\unit{mm}.

An outline of the tested sensors' characteristics, as well as the used parameters is given in table~\ref{tab:sensors}.

\begin{figure*}[ht!]
	\centering%
	\includegraphics[width=0.49\linewidth]{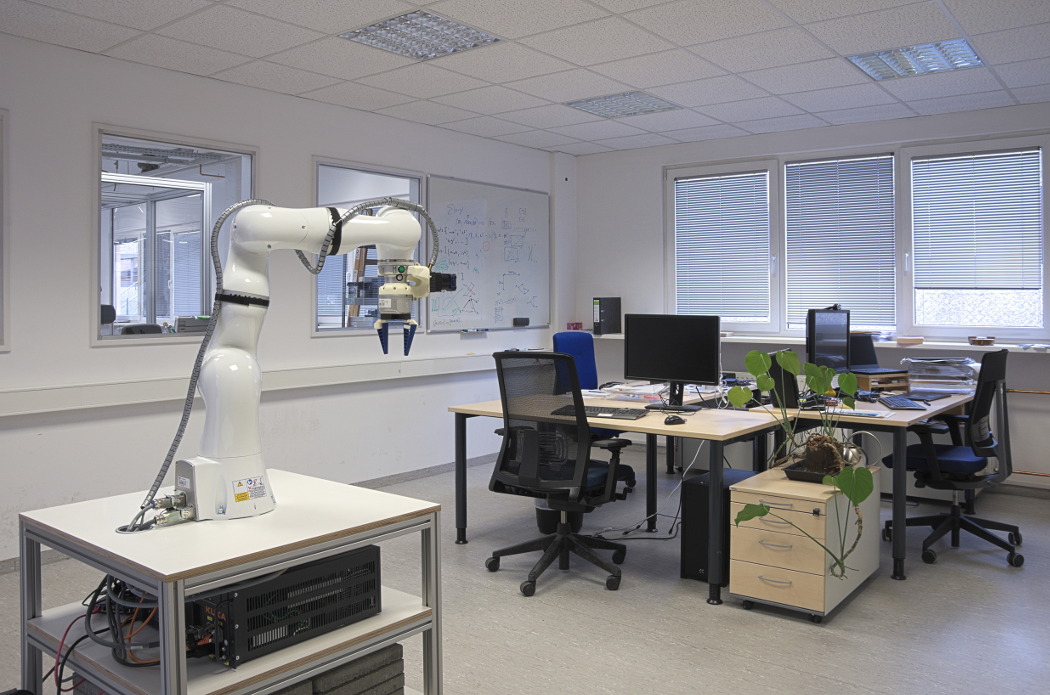}\hfill%
	\includegraphics[width=0.49\linewidth]{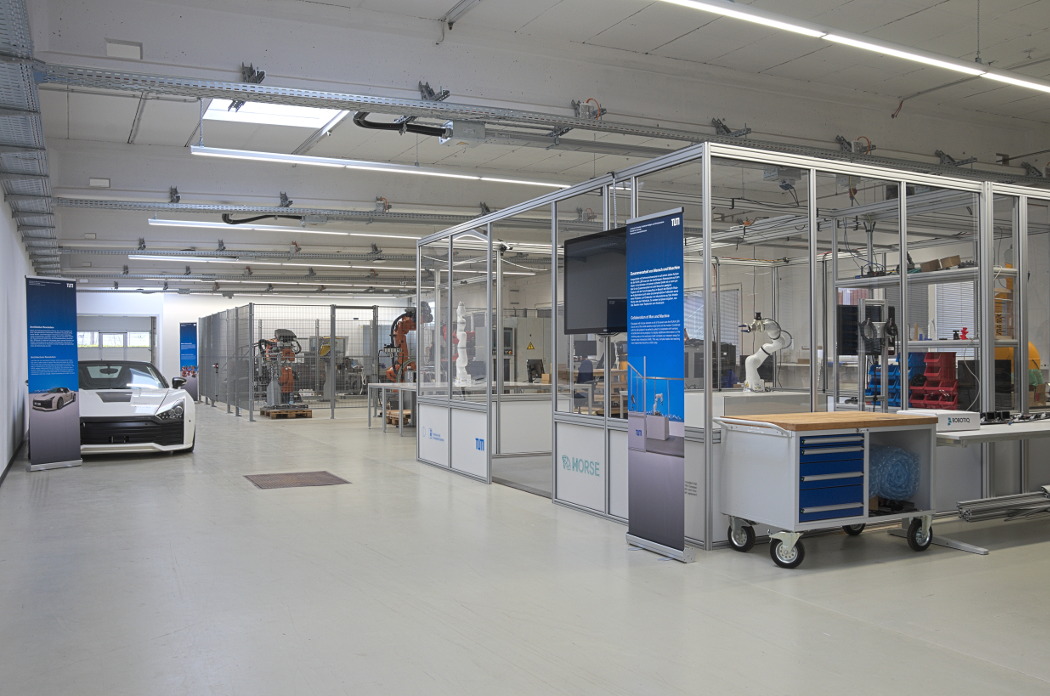}\\%
	\vspace{0.02\linewidth}%
	\includegraphics[width=0.49\linewidth]{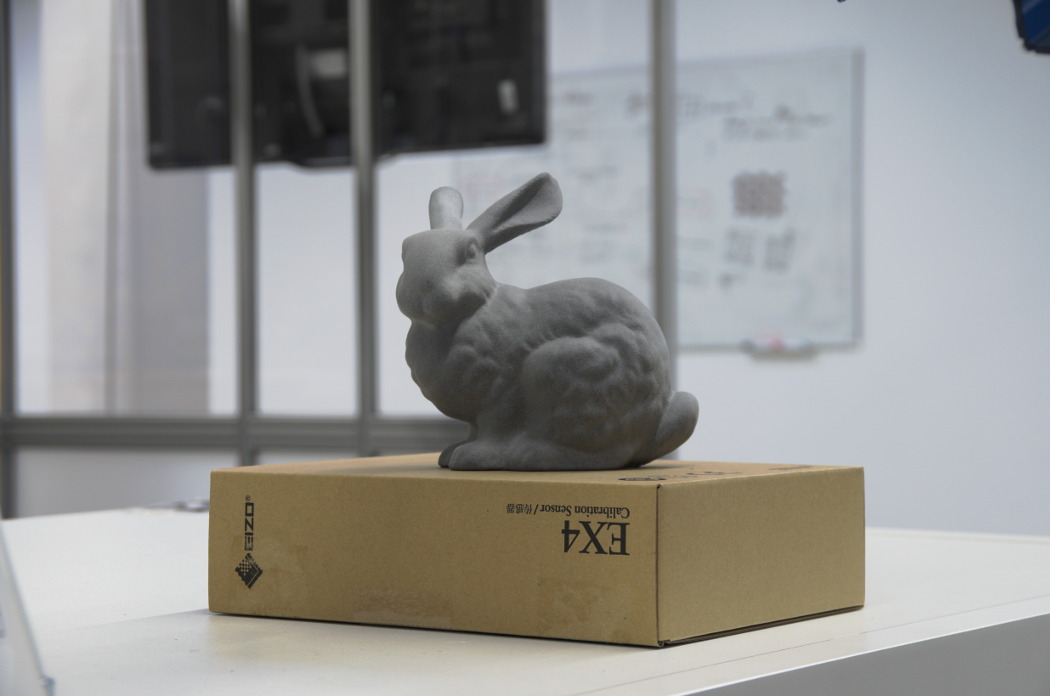}\hfill%
	\includegraphics[width=0.49\linewidth]{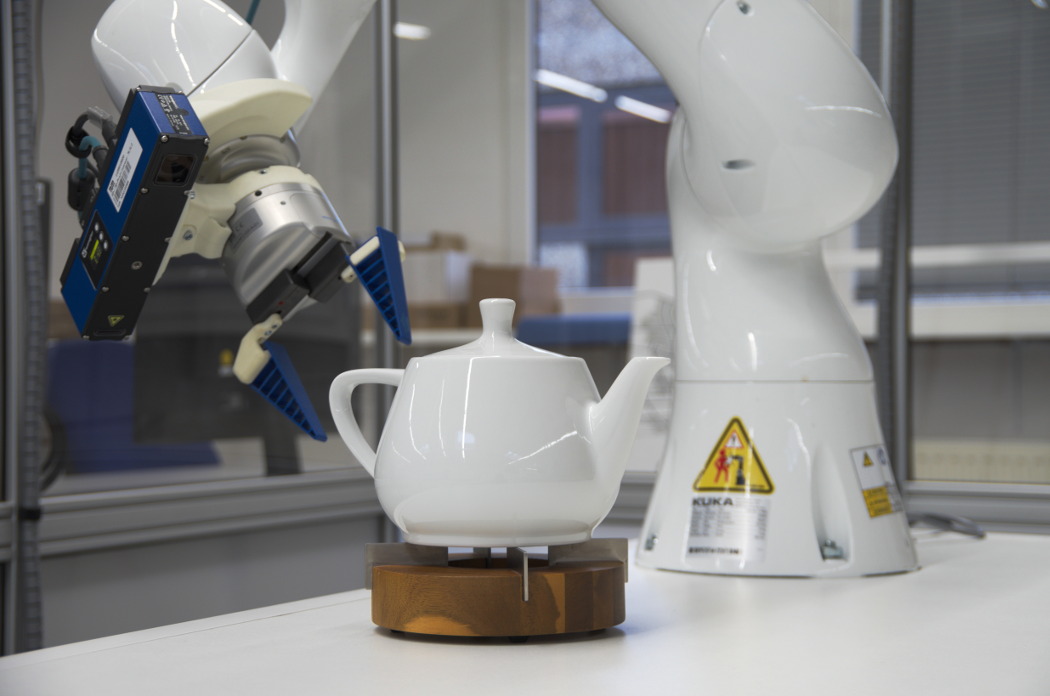}\\%
	\caption{
		Pictures of the scanned real world scenes. 1) Office, 2) TUM laboratory (the robot is positioned is in the workcell on the right), 3) Standford Bunny and 4) Utah Teapot (1.4\unit{l} version).
	}%
	\label{fig:scenes}
\end{figure*}

The evaluation is performed on two close range scenes of basic objects---a 3D printed Stanford Bunny (originally scanned and made public by Turk and Levoy using their back then new scanning algorithm \cite{turk1994zippered}) and a Utah teapot (also known as Newell Teapot) \cite{newell1975utilization, crow1987origins} which is still in production by its original manufacturer---as well as two large scale scenes, featuring an office and a laboratory at TUM (see figure~\ref{fig:scenes}).

The Hokuyo LiDAR has an extensive reach and a view angle of {270\degree} making its orientation neglectable as it is almost always capable of seeing a surface somewhere within any indoor scene.
Thus recordings were performed by moving the robot to seven randomly generated configurations and moving one of it's axes by 180\degree, each at a time.
The process was repeated five times, resulting in four batches of with a total of 28 scans and one validation set with seven scans.
The joint velocities were chosen in a way to obtain a similar density between scan lines and points within a single line.

For the other used sensors the devices' fields of view need to be taken into account when selecting the scanning poses, to assure sufficient overlap in the projected point clouds of the datasets.
In these recordings the scanning poses were chosen by manually selecting a suitable sensor placements and using an inverse kinematics solver to find a matching robot configuration for the desired EE poses.
In case of the Wenglor MLSL236 the trajectories were defined by a fixed start and end configuration in joint space.
The robot was moved between those configurations with a constant velocity to allow linear interpolation of the joint positions.
For the two depth cameras the robot was moved to fixed positions from which only single images were taken.
Figure~\ref{fig:joint_positions} provides an detailed overview of the selected configurations and trajectories.

\begin{table*}[p]
	\caption{Overview of calibrated parameters}%
	\label{tab:calibrated_parameters}%
	\vspace{-1mm}%
	\newcommand*{\urdf}{$\medcircle$}%
	\newcommand*{\calib}{$\medblackcircle$}%
	\newcommand*{\vicon}{$\medblackdiamond$}%
	\setlength{\tabcolsep}{0.225em}%
	\centering%
	\begin{tabular}{l | cccccc | cccc | cccc | cccc | cccc | cccc | cccc | cccccc}
		\toprule
		& \multicolumn{6}{c|}{$\mathcal{O}$ to $\jf{1}$} & \multicolumn{4}{c|}{$\jf{1}$ to $\jf{2}$} & \multicolumn{4}{c|}{$\jf{2}$ to $\jf{3}$} & \multicolumn{4}{c|}{$\jf{3}$ to $\jf{4}$} & \multicolumn{4}{c|}{$\jf{4}$ to $\jf{5}$} & \multicolumn{4}{c|}{$\jf{5}$ to $\jf{6}$} & \multicolumn{4}{c|}{$\jf{6}$ to $\jf{7}$} & \multicolumn{6}{c}{$\jf{7}$ to $\mathcal{E}$} \\ 
		& $\alpha$ & $\beta$ & $\gamma$ & $x$ & $y$ & $z$ & $\alpha$ & $\beta$ & $x$ & $y$ & $\alpha$ & $\beta$ & $x$ & $y$  & $\alpha$ & $\beta$ & $x$ & $y$  & $\alpha$ & $\beta$ & $x$ & $y$ & $\alpha$ & $\beta$ & $x$ & $y$ & $\alpha$ & $\beta$ & $x$ & $y$ & $\alpha$ & $\beta$ & $\gamma$ & $x$ & $y$ & $z$ \\
		\midrule
		Calibratable Parameters & - & - & - & - & - & - & \calib & - & \calib & - & \calib & \calib & \calib & \calib & \calib & \calib & \calib & \calib & \calib & \calib & \calib & \calib & \calib & \calib & \calib & \calib & \calib & \calib & \calib & \calib & \calib & \calib & \calib & \calib & \calib & \calib \\
		Model for Comparison & \vicon & \vicon & \vicon & \vicon & \vicon & \vicon & \calib & \vicon & \calib & \vicon & \calib & \calib & \calib & \calib & \calib & \calib & \calib & \calib & \calib & \calib & \calib & \calib & \calib & \calib & \calib & \calib & \calib & \calib & \calib & \calib & \vicon & \vicon & \vicon & \vicon & \vicon & \vicon \\
		\midrule
		URDF Model & \urdf & \urdf & \urdf & \urdf & \urdf & \urdf & \urdf & \urdf & \urdf & \urdf & \urdf & \urdf & \urdf & \urdf & \urdf & \urdf & \urdf & \urdf & \urdf & \urdf & \urdf & \urdf & \urdf & \urdf & \urdf & \urdf & \urdf & \urdf & \urdf & \urdf & \urdf & \urdf & \urdf & \urdf & \urdf & \urdf \\
		URDF + Calibrated Origin & \vicon & \vicon & \vicon & \vicon & \vicon & \vicon & \urdf & \vicon & \urdf & \vicon & \urdf & \urdf & \urdf & \urdf & \urdf & \urdf & \urdf & \urdf & \urdf & \urdf & \urdf & \urdf & \urdf & \urdf & \urdf & \urdf & \urdf & \urdf & \urdf & \urdf & \urdf & \urdf & \urdf & \urdf & \urdf & \urdf \\
		URDF + Calibrated Origin \& EE & \vicon & \vicon & \vicon & \vicon & \vicon & \vicon & \urdf & \vicon & \urdf & \vicon & \urdf & \urdf & \urdf & \urdf & \urdf & \urdf & \urdf & \urdf & \urdf & \urdf & \urdf & \urdf & \urdf & \urdf & \urdf & \urdf & \urdf & \urdf & \urdf & \urdf & \vicon & \vicon & \vicon & \vicon & \vicon & \vicon \\
		Traditional Calibration & \vicon & \vicon & \vicon & \vicon & \vicon & \vicon & \vicon & \vicon & \vicon & \vicon & \vicon & \vicon & \vicon & \vicon & \vicon & \vicon & \vicon & \vicon & \vicon & \vicon & \vicon & \vicon & \vicon & \vicon & \vicon & \vicon & \vicon & \vicon & \vicon & \vicon & \vicon & \vicon & \vicon & \vicon & \vicon & \vicon \\
		\bottomrule
	\end{tabular}
	\vspace{0.1em}\\
	\urdf\hspace{0.4em}Parameters from CAD model / manual measuring%
	\hspace{2em}\calib\hspace{0.4em}Calibrated using presented approach%
	\hspace{2em}\vicon\hspace{0.4em}Calibrated using optical tracking system%
	\renewcommand\theadalign{bc}%
	\renewcommand\theadfont{\bfseries}%
	\renewcommand\theadgape{\Gape[4pt]}%
	\renewcommand\cellgape{\Gape[4pt]}%
	\vspace{3em}

	\caption{Comparison of results to traditional calibration}%
	\label{tab:calibration_results}
	\vspace{-1mm}%
	\setlength{\tabcolsep}{0.4em}%
	\centering%
	\begin{tabular}{llr rrrrrrrr}
		\toprule
		 & \thead{Dataset} & \thead{Scans} & \thead{Orientation\\Error [deg]} & \thead{Position\\Error [mm]} & \thead{Outliers} & \thead{Filtered\\Orientation\\Error [deg]} & \thead{Filtered\\Position\\Error [mm]} & \thead{Iterations} & \thead{Valid Point\\Matches per\\Iteration} & \thead{Runtime [s]}\\
		 \midrule
		 \multirow{8}{*}{\rotatebox[origin=c]{90}{\parbox{2cm}{\centering Hokuyo\\UTM-30LX}}}
		   & \multirow{4}{*}{Lab}
		   &  7 & 1.287 & 25.71 &  25 & 0.744 & 22.95 & 50 &  1\,190\,814.7 &  5573.0 \\
		 & & 14 & 1.178 & 10.45 &  23 & 0.636 &  7.39 & 16 &  5\,411\,943.3 &  8129.9 \\
		 & & 21 & 1.148 & 10.68 &  23 & 0.604 &  7.59 &  8 & 11\,413\,095.6 & 11076.0 \\
		 & & 28 & 1.114 & 11.76 &  23 & 0.567 &  8.77 &  7 & 21\,016\,877.9 & 18932.5 \\
		 \cmidrule{2-11}
		   & \multirow{4}{*}{Office}
		   &  7 & 1.217 & 18.80 &  24 & 0.675 & 15.87 & 50 &  1\,970\,957.2 &  9997.9 \\
		 & & 14 & 1.103 &  7.94 &  23 & 0.556 &  4.80 &  9 &  9\,062\,888.7 &  8835.3 \\
		 & & 21 & 1.112 &  7.65 &  23 & 0.565 &  4.48 & 14 & 19\,224\,321.9 & 25507.3 \\
		 & & 28 & 1.115 &  7.25 &  23 & 0.569 &  4.06 &  7 & 35\,045\,811.3 & 27312.3 \\
		 \midrule
		 \multirow{8}{*}{\rotatebox[origin=c]{90}{\parbox{2cm}{\centering Wenglor\\MLSL236}}}
		   & \multirow{4}{*}{Bunny}
		   &  7 & 1.267 &  7.62 &  23 & 0.727 &  4.45 & 50 &  1\,092\,781.1 &  6627.7 \\
		 & & 14 & 1.102 &  5.08 &  23 & 0.555 &  1.80 & 17 &  5\,979\,392.7 &  9014.5 \\
		 & & 21 & 1.108 &  5.04 &  23 & 0.562 &  1.76 &  7 & 12\,093\,808.4 &  9182.5 \\
		 & & 28 & 1.108 &  5.15 &  23 & 0.561 &  1.87 &  7 & 21\,902\,965.6 & 16269.6 \\
		 \cmidrule{2-11}
		   & \multirow{4}{*}{Teapot}
		   &  7 & 1.138 &  5.77 &  23 & 0.592 &  2.51 & 50 &  1\,408\,625.3 &  7617.3 \\
		 & & 14 & 1.100 &  5.47 &  23 & 0.553 &  2.19 &  7 &  7\,820\,361.6 &  5564.8 \\
		 & & 21 & 1.116 &  5.32 &  23 & 0.569 &  2.04 &  8 & 17\,363\,749.6 & 13996.8 \\
		 & & 28 & 1.116 &  5.24 &  23 & 0.571 &  1.96 &  6 & 33\,530\,538.7 & 21414.3 \\
		 \midrule
		 \multirow{16}{*}{\rotatebox[origin=c]{90}{\parbox{4.2cm}{\centering Microsoft Kinect Azure}}}
		   & \multirow{4}{*}{Lab}
		   &  7 & 1.299 & 12.73 &  24 & 0.757 &  9.84 & 12 &     140\,739.9 &    97.8 \\
		 & & 14 & 1.111 &  6.74 &  23 & 0.564 &  3.55 & 12 &     239\,020.8 &   186.1 \\
		 & & 21 & 1.104 &  7.27 &  23 & 0.556 &  4.07 & 12 &     382\,920.5 &   333.1 \\
		 & & 28 & 1.101 &  6.79 &  23 & 0.552 &  3.57 & 16 &     724\,856.7 &   819.4 \\
		 \cmidrule{2-11}
		   & \multirow{4}{*}{Office}
		   &  7 & 1.364 & 59.93 & 329 & 0.822 & 35.13 & 18 &     249\,166.5 &   317.2 \\
		 & & 14 & 1.103 &  7.33 &  23 & 0.555 &  4.13 &  8 &     540\,193.1 &   324.2 \\
		 & & 21 & 1.108 &  6.35 &  23 & 0.561 &  3.10 &  8 &  1\,408\,679.1 &   642.3 \\
		 & & 28 & 1.102 &  5.78 &  23 & 0.553 &  2.50 &  7 &  2\,133\,716.0 &   684.7 \\
		 \cmidrule{2-11}
		   & \multirow{4}{*}{Bunny}
		   &  7 & 1.566 & 27.33 &  24 & 1.037 & 14.42 & 16 &     576\,383.5 &   444.4 \\
		 & & 14 & 1.172 & 10.48 &  23 & 0.626 &  7.43 & 10 &  2\,452\,292.9 &  1256.3 \\
		 & & 21 & 1.116 &  7.15 &  23 & 0.567 &  3.97 &  8 &  5\,009\,795.9 &  2135.6 \\
		 & & 28 & 1.114 &  6.50 &  23 & 0.566 &  3.28 &  9 &  8\,736\,878.1 &  4226.9 \\
		 \cmidrule{2-11}
		   & \multirow{4}{*}{Teapot}
		   &  7 & 1.311 & 10.17 &  24 & 0.770 &  7.03 &  8 &     590\,497.3 &   257.4 \\
		 & & 14 & 1.142 &  7.72 &  23 & 0.594 &  4.57 &  8 &  2\,465\,504.3 &  1082.6 \\
		 & & 21 & 1.103 &  5.91 &  23 & 0.555 &  2.67 &  7 &  5\,184\,669.3 &  2065.8 \\
		 & & 28 & 1.103 &  5.34 &  23 & 0.556 &  2.07 &  6 &  9\,190\,931.3 &  3228.6 \\
		 \midrule
		 \multirow{8}{*}{\rotatebox[origin=c]{90}{\parbox{2cm}{\centering PhotoNeo MotionCam 3D}}}
		   & \multirow{4}{*}{Bunny}
		   &  7 & 1.190 & 18.48 &  25 & 0.642 & 15.60 & 38 &  1\,666\,129.5 &  3639.1 \\
		 & & 14 & 1.109 &  5.48 &  23 & 0.562 &  2.21 & 18 &  6\,206\,851.7 &  6034.9 \\
		 & & 21 & 1.101 &  5.11 &  23 & 0.553 &  1.82 & 14 & 14\,566\,960.8 & 11531.8 \\
		 & & 28 & 1.094 &  4.99 &  23 & 0.546 &  1.70 & 11 & 22\,572\,019.6 & 15984.8 \\
		 \cmidrule{2-11}
		   & \multirow{4}{*}{Teapot}
		   &  7 & 1.166 & 11.31 &  23 & 0.622 &  8.27 & 47 &  1\,511\,654.7 &  3951.3 \\
		 & & 14 & 1.140 &  5.68 &  23 & 0.595 &  2.41 & 35 &  4\,468\,177.3 &  8495.6 \\
		 & & 21 & 1.108 &  5.25 &  23 & 0.561 &  1.97 & 18 & 10\,240\,472.9 & 10479.5 \\
		 & & 28 & 1.096 &  5.10 &  23 & 0.548 &  1.82 & 15 & 14\,713\,290.9 & 12979.9 \\
		 \midrule
		 \multirow{4}{*}{\rotatebox[origin=c]{90}{\parbox{1.25cm}{\centering Traditional\\Calibration}}}
		 & URDF     &    - & 2.057 & 35.90 & 73 & 1.522 & 35.32 & & \\
		 & + Origin & 4500 & 2.016 & 24.46 & 27 & 1.484 & 21.91 & & \\
		 & + EE     & 4500 & 1.103 &  6.24 & 23 & 0.556 &  3.00 & & \\
		 & Full     & 4500 & 1.095 &  5.05 & 23 & 0.547 &  1.77 & & \\
		\bottomrule
	\end{tabular}
\end{table*}

\begin{figure*}[htbp]%
	\centering%
	\newlength{\cloudImageWidth}%
	\setlength{\cloudImageWidth}{0.45\textwidth}%
	\begin{tabular}{cc}
		\begin{minipage}{0.46\textwidth}\includegraphics[width=\linewidth]{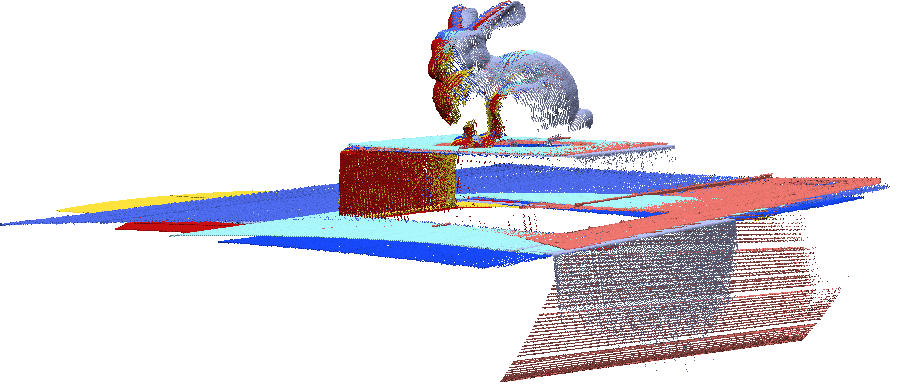}\end{minipage}&
		\begin{minipage}{0.46\textwidth}\includegraphics[width=\linewidth]{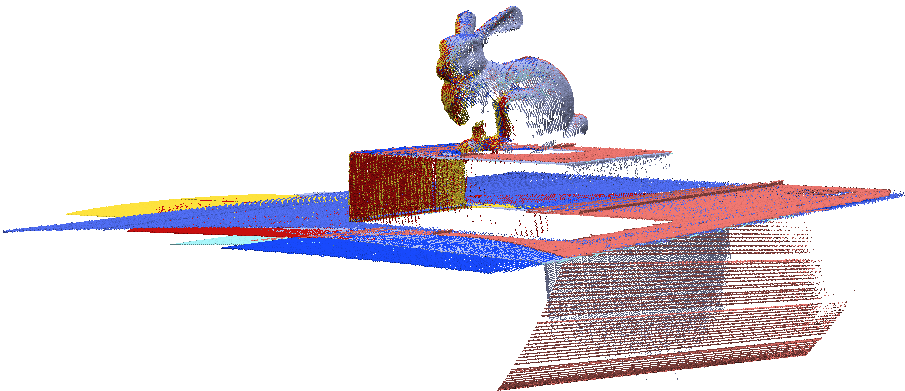}\end{minipage}\\
		\addlinespace[0.25em]
		\begin{minipage}{\cloudImageWidth}\includegraphics[width=\linewidth]{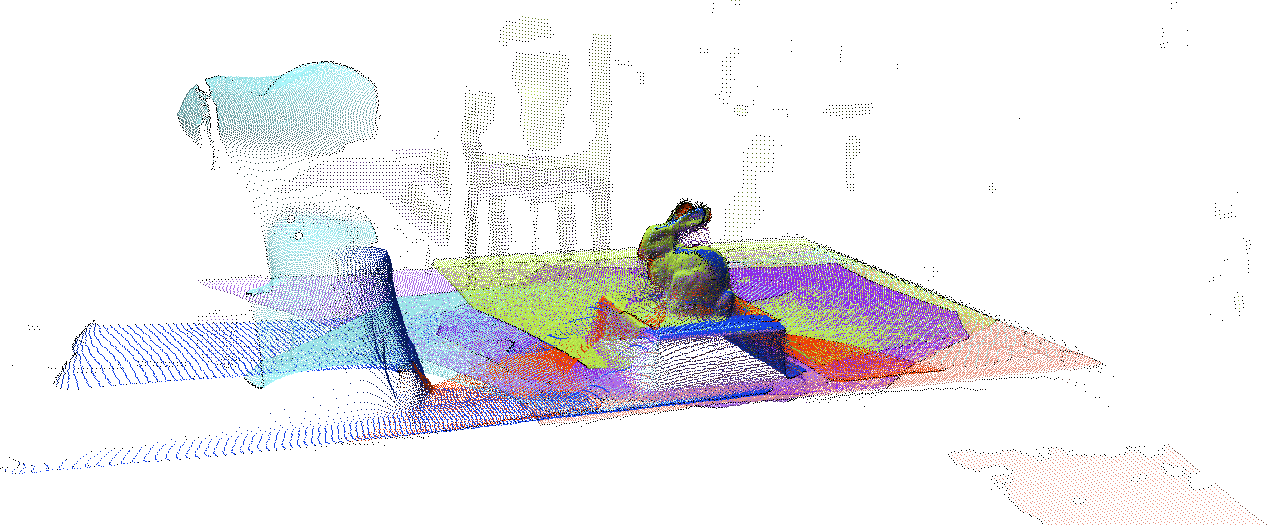}\end{minipage}&
		\begin{minipage}{\cloudImageWidth}\includegraphics[width=\linewidth]{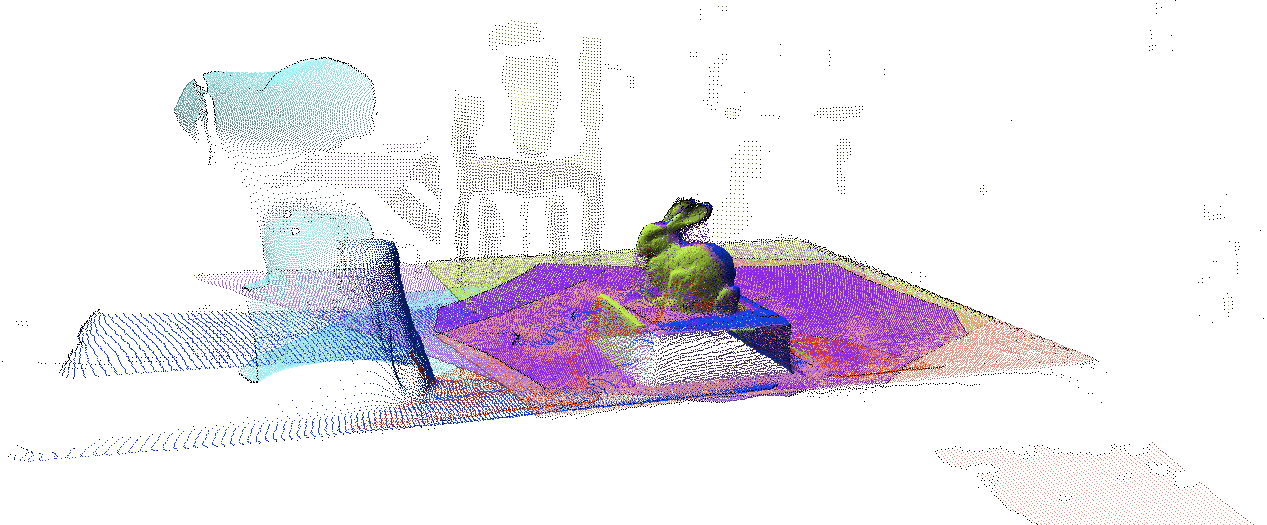}\end{minipage}\\
		\addlinespace[-1em]
		\begin{minipage}{\cloudImageWidth}\includegraphics[width=\linewidth]{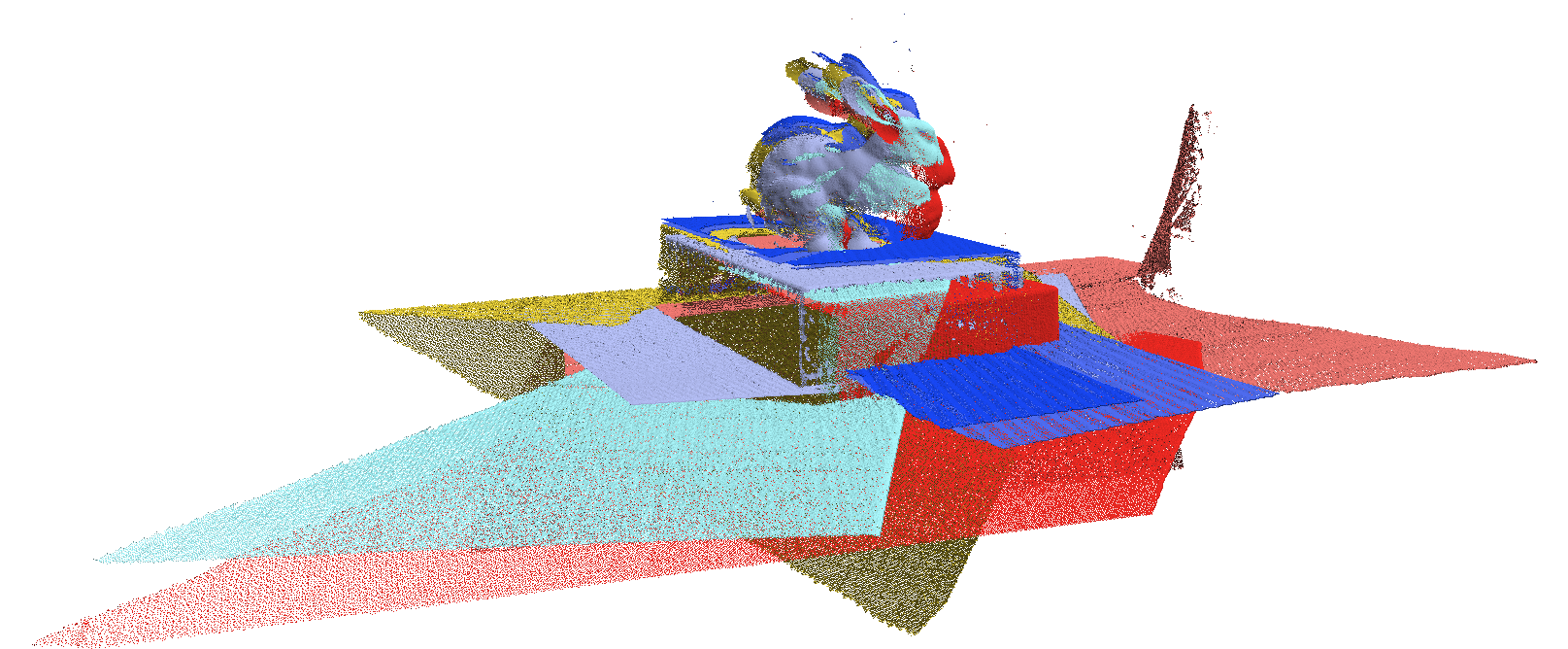}\end{minipage}&
		\begin{minipage}{\cloudImageWidth}\includegraphics[width=\linewidth]{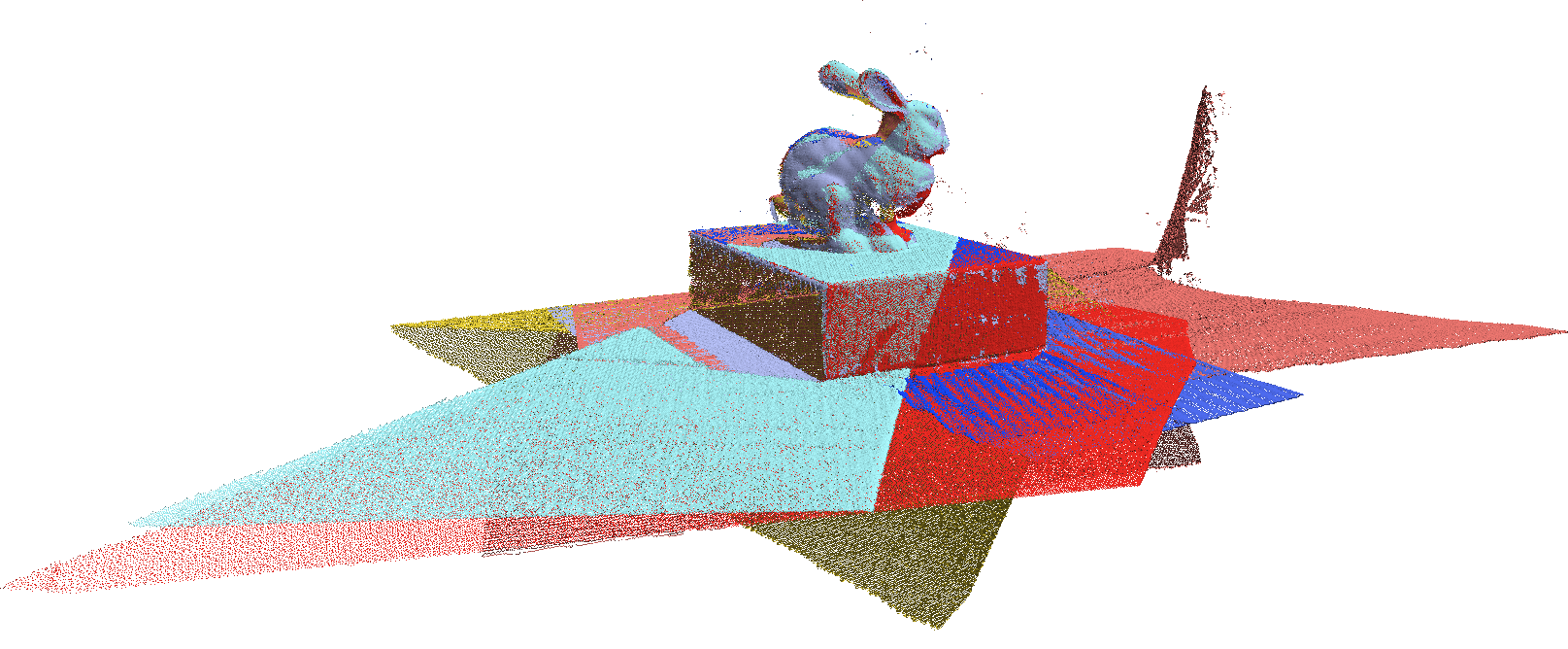}\end{minipage}\\
		\addlinespace[2em]
		\begin{minipage}{0.40\textwidth}\includegraphics[width=\linewidth]{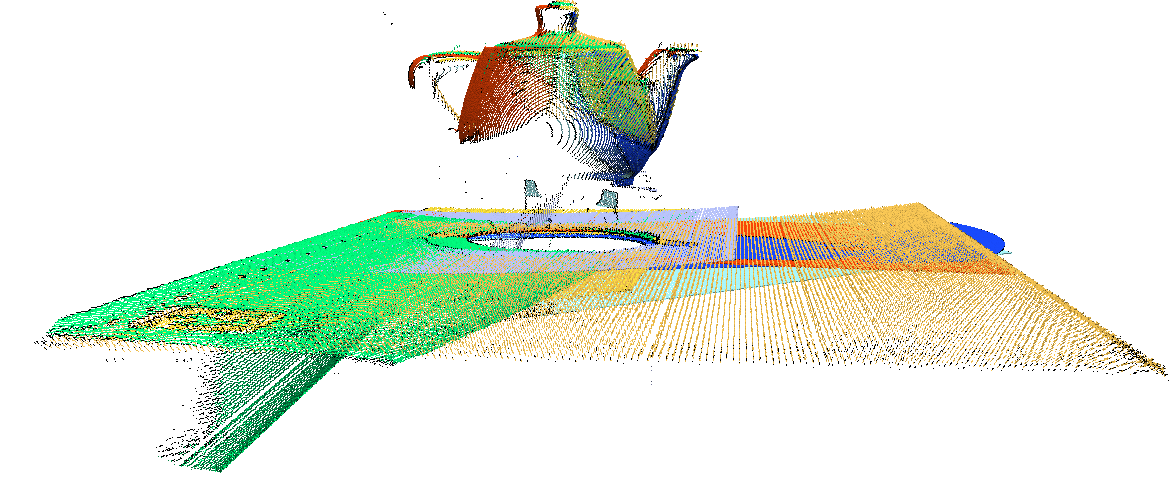}\end{minipage}&
		\begin{minipage}{0.40\textwidth}\includegraphics[width=\linewidth]{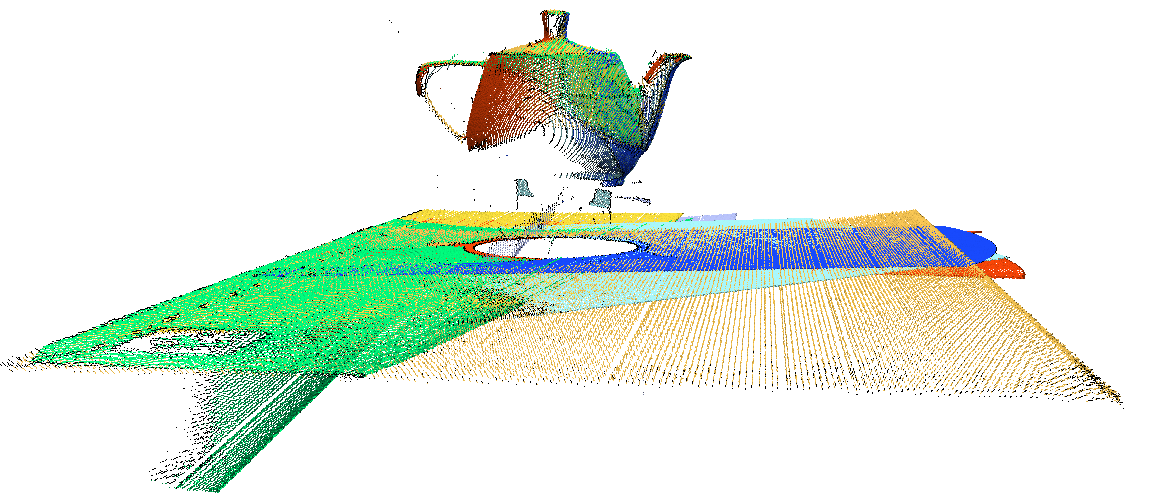}\end{minipage}\\
		\begin{minipage}{\cloudImageWidth}\includegraphics[width=\linewidth]{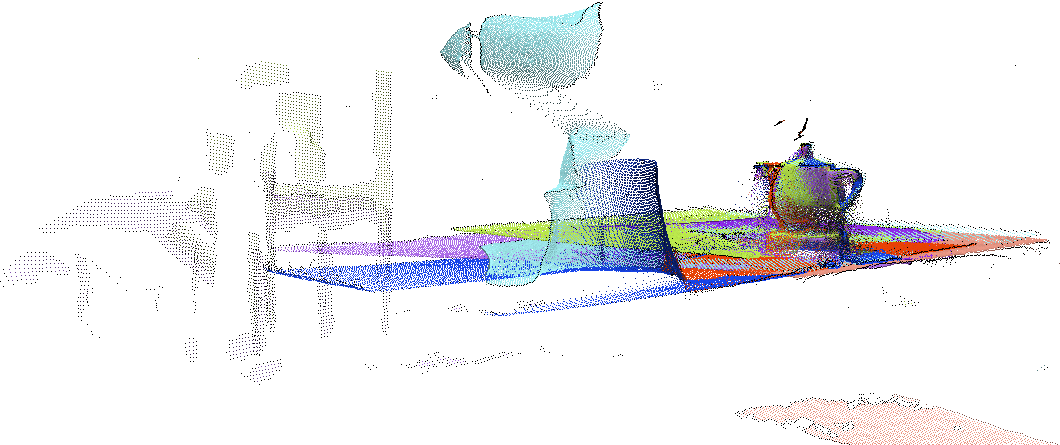}\end{minipage}&
		\begin{minipage}{\cloudImageWidth}\includegraphics[width=\linewidth]{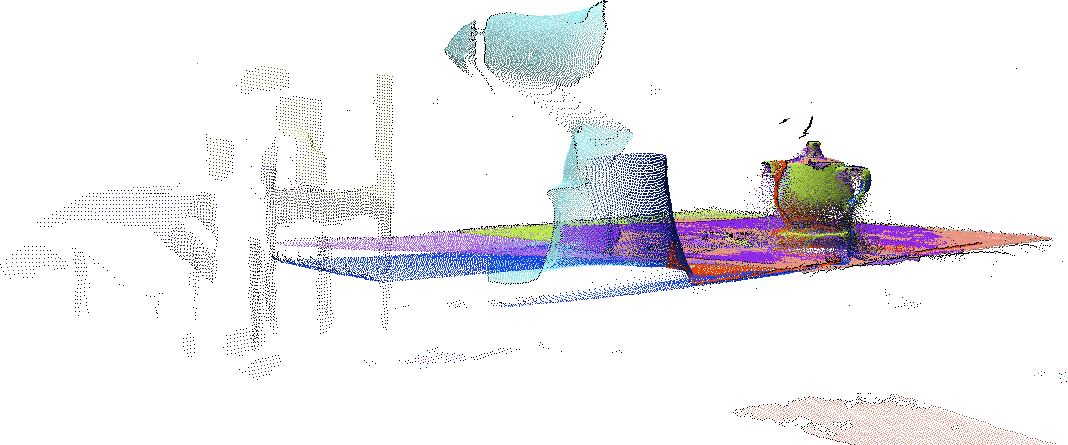}\end{minipage}\\
		\addlinespace[-0.5em]
		\begin{minipage}{0.28\linewidth}\includegraphics[width=\linewidth]{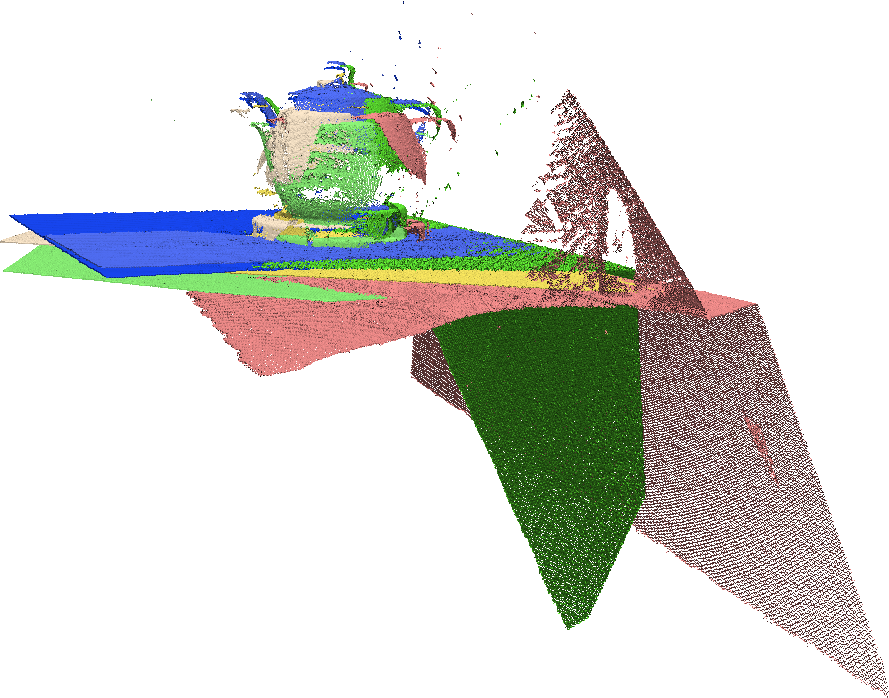}\end{minipage}&
		\begin{minipage}{0.28\linewidth}\includegraphics[width=\linewidth]{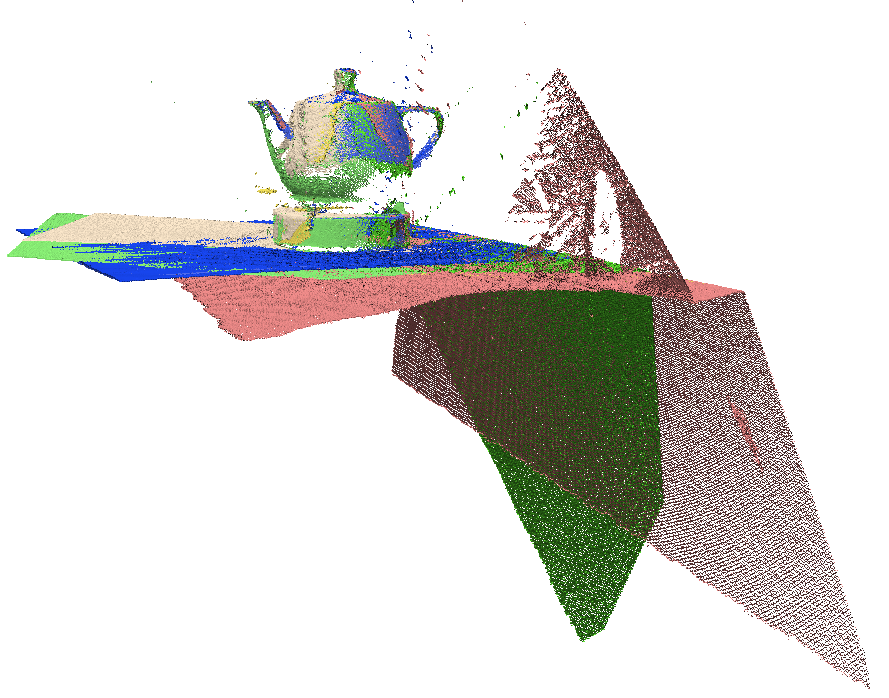}\end{minipage}\\
		\addlinespace[0.5em]
	\end{tabular}
	\caption{
		Point clouds projected by uncalibrated (left) and calibrated systems (right) on 28 datasets.
		Each pictures shows seven validation datasets that were not used for the calibration itself.
		From top to bottom:
		Stanford Bunny recorded with 1) Wenglor MLSL236, 2) Kinect Azure, and 3) Photoneo MotionCam,
		followed by Utah Teapot recorded with 4) Wenglor MLSL236, 5) Kinect Azure and 6) Photoneo MotionCam.
		The wooden warmer below the Teapot is not visible to the UV laser of the Wenglor scanner.
		Differences in the alignment of the point clouds of the Kinect Azure are best noticeable on the partly visible background structures.
	}%
	\label{fig:pointclouds1}
\end{figure*}

\begin{figure*}[htbp]
	\centering%
	\begin{tabular}{cc}
		\begin{minipage}{0.45\textwidth}\includegraphics[width=\linewidth]{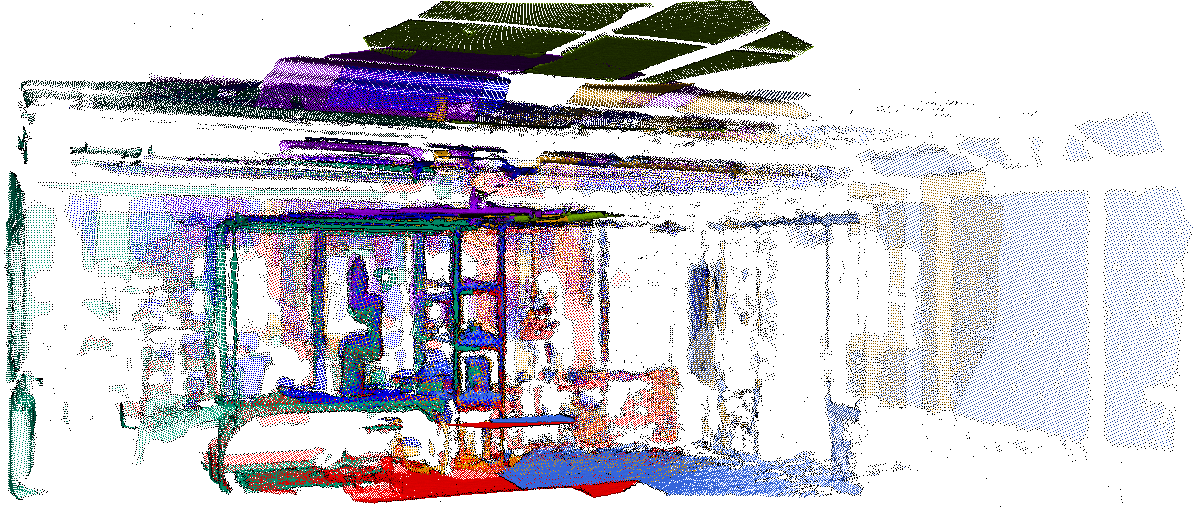}\end{minipage}&
		\begin{minipage}{0.45\textwidth}\includegraphics[width=\linewidth]{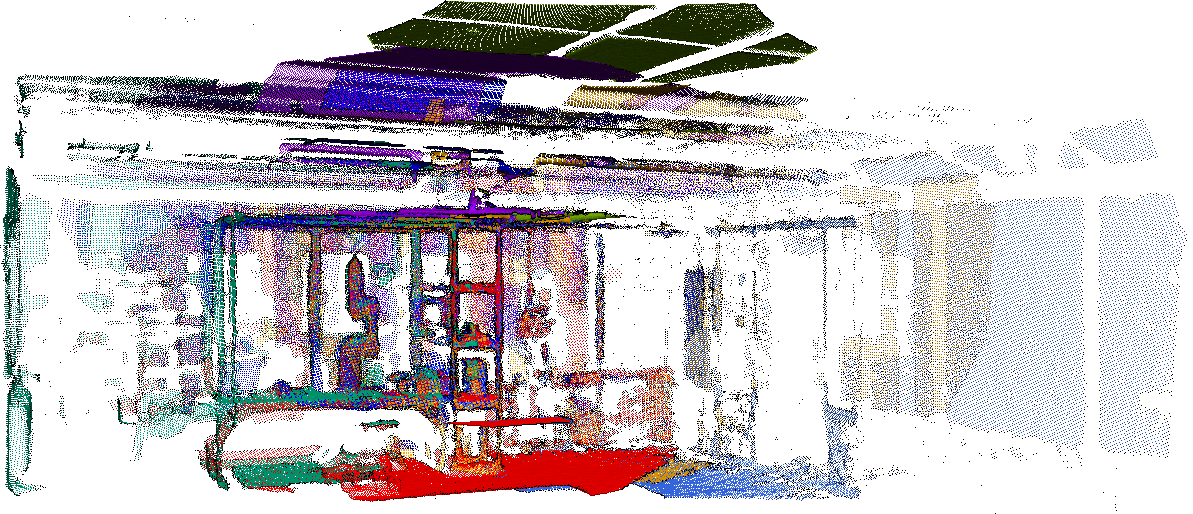}\end{minipage}\\
		\addlinespace[1.0em]
		\begin{minipage}{0.48\textwidth}\includegraphics[width=\linewidth]{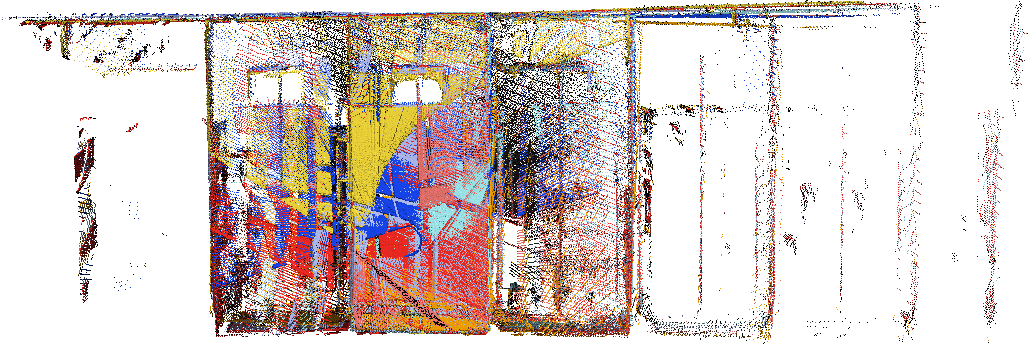}\end{minipage}&
		\begin{minipage}{0.48\textwidth}\includegraphics[width=\linewidth]{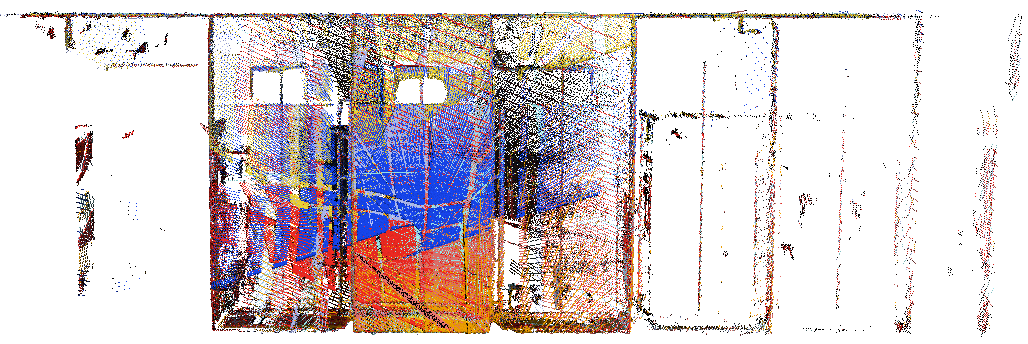}\end{minipage}\\
		\addlinespace[2em]
		\begin{minipage}{0.48\textwidth}\includegraphics[width=\linewidth]{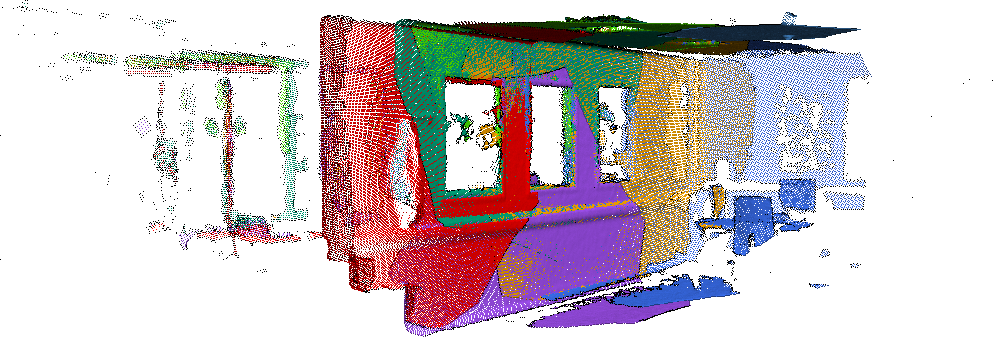}\end{minipage}&
		\begin{minipage}{0.48\textwidth}\includegraphics[width=\linewidth]{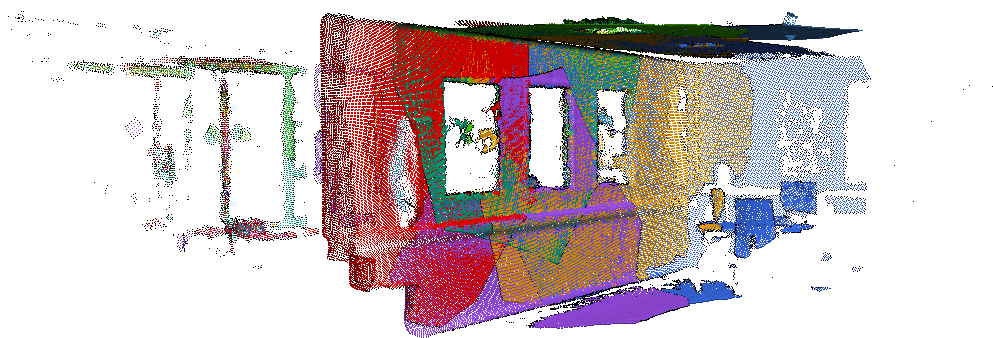}\end{minipage}\\
		\addlinespace[0.75em]
		\begin{minipage}{0.45\textwidth}\includegraphics[width=\linewidth]{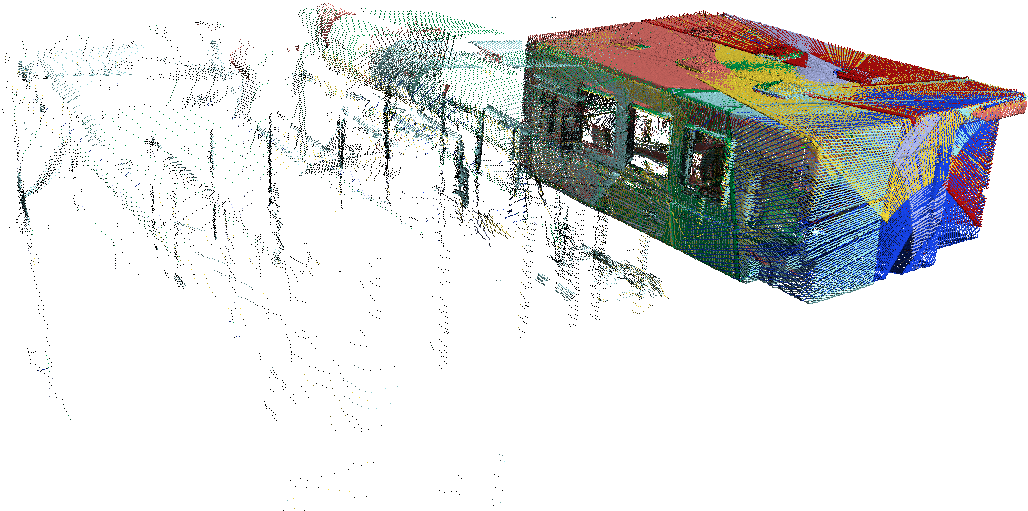}\end{minipage}&
		\begin{minipage}{0.45\textwidth}\includegraphics[width=\linewidth]{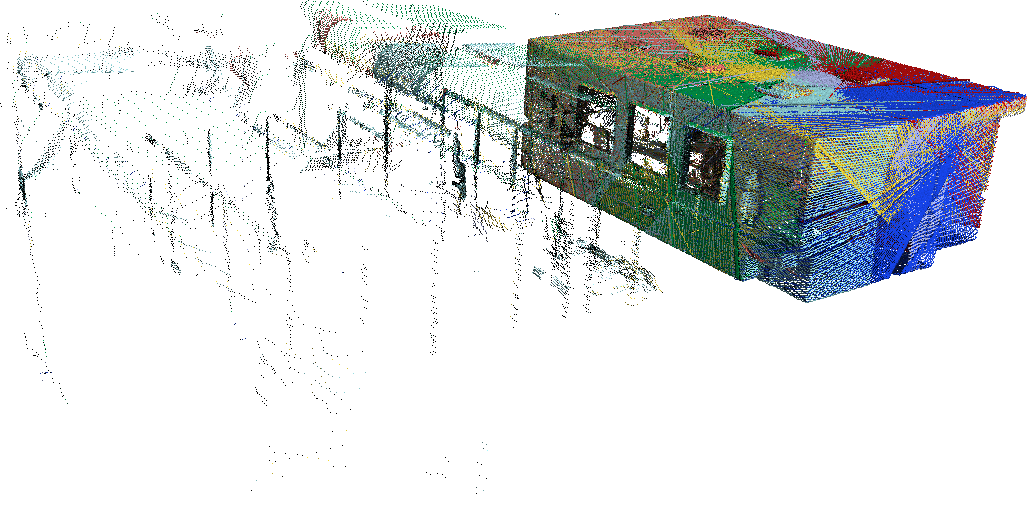}\end{minipage}\\
		\addlinespace[-0.75em]
	\end{tabular}
	\caption{
		Point clouds projected by uncalibrated (left) and calibrated systems (right) on 28 datasets.
		Each pictures shows seven validation datasets that were not used for the calibration itself.
		From top to bottom:
		TUM laboratory recorded with 1) Kinect Azure, and 2) Hokuyo UTM-30LX (viewed from top),
		followed by an office recorded with 3) Kinect Azure, and 4) Hokuyo UTM-30LX.
	}%
	\label{fig:pointclouds2}
\end{figure*}

A reference calibration was obtained using an optical tracking system based on five Vicon Vero v1.3 cameras.
For this calibration a marker was placed on the ground next to the robot's based while a second one was mounted to the EE.
The robot was then moved to 5\,000 randomly selected configurations while the poses of both markers as well as the robot's joint positions were recorded.
90\unit{\%} of those points were used to compute the optimal MCPC parameters connecting the static marker next to the robot's base $\mathcal{O}$ with the one attached to the end-effector $\mathcal{E}$.
Based on the found model, outliers in the measurements with an offset of more than 5\unit{cm} and/or 0.05\unit{rad} (2.86\degree) from the estimated EE pose were excluded from the recorded data.
The procedure was repeated five times.
Based on the remaining 500 poses the reference calibration reaches an position error of 1.77\unit{mm} and an orientation error of {0.547\degree}, when applying the same outlier filtering on these measurements as well.

The initial model for both calibration attempts was obtained from CAD data of the used components whenever available.
Parts for which such data was not accessible were measured manually.

As the presented approach takes the coordinate frame of the used 3D sensor as an endpoint of the kinematic chain, only the estimated MCPC parameters of the robot itself are comparable to the reference calibration.
To enable an evaluation on the same 500 validation measurements, the found transformation between the seventh joint $\mathcal{J}_7$ and $\mathcal{E}$ is thus replaced by the reference one.
The same applies for the transformation from $\mathcal{O}$ and $\mathcal{J}_1$ as well as the non-calibratable parameters between $\mathcal{J}_1$ and $\mathcal{J}_2$.
An overview of the calibrated parameters by the different approaches is given in table \ref{tab:calibrated_parameters}.
To ensure similar conditions the outlier filtering on the verification data is applied here as well.

For the final results, the maximum number of iterations $i_\idx{max}$ was limited to 50.
All retrieved calibration results are listed in table~\ref{tab:calibration_results}.
The runtime measurements were performed on a workstation PC equipped with an AMD Ryzen 9 5950X CPU (2020 model with 16 physical cores) and 128GB RAM, running a multi-threaded C++ implementation of the described framework under Ubuntu 18.04 LTS.
Figures~\ref{fig:pointclouds1} and \ref{fig:pointclouds2} further show the captured validation sets to allow a subjective impression of the achieved calibration quality.

The presented results allow for a number of observations and conclusions:
As one can see in table~\ref{tab:calibration_results}, using only seven scans is usually insufficient to reach a suitable calibration result.
Either the precision is way worse than what can be expected from a higher number of scans, or the ICP did not converge at all.
There is also an clear trend towards reaching a higher precision when using more data.

Also, the findings confirm an expected relation between the the calibration result is the precision of the used sensor.
For high precision sensors such as the Wenglor MLSL236 or the PhotoNeo MotionCam 3D the results are similar to the ones obtained by the tracking system.
In case of the MotionCam 3D on the Stanford Bunny scene the proposed framework even found a solution which is slightly better than the reference.
However, even with a Kinect Azure one may obtain results almost as good a as reference, even though its cost are less the one fiftieth of the tracking system's one.

The observations further raise the suspicion that less complex, small-scale scenes are better suited for calibration.
Especially in combination of large scenes captured by far range sensors even smallest orientation errors have a strong effect on the overall error metric.
On the example of the lab scan taken with the Hokuyo LiDAR one can observe that the orientation error is continuously reduced with an increasing number of used datasets, even at the expense of the position error.

Evidently the runtime is directly dependent on the number of used iterations and found point matches---which is again related to the number of scans.
Fortunately one can see, that the number of required iterations goes down when the overall number of datasets increases.

Finally, while recording the used datasets I made another observation:
The spatial position of the sensor is---except for ensuring sufficient overlap in the scan data---neglectable.
It is, however, desirable to reach a complete and uniform sampling of the manipulator's joint space.
Especially extrapolations in joint ranges not covered by the calibration data will result in errors.
When closely studying figure~\ref{fig:joint_positions} one can see a direct relation between the coverage of the joint space and the calibration precision.

\section{Summary}

A new framework for fully autonomous calibration of robot manipulators has been presented, by extending the ICP algorithm to allow optimization of complex kinematic models instead of estimating a single, rigid transformation.
The shown implementation has been evaluated on multiple real world scenes on various hardware configurations.
Comparing the results to a dedicated tracking system has clearly shown the functionality of the framework.
More than that: The achieved precision is similar to the one of the way more expensive reference system.
Given the right scene, even with a Microsoft Kinect Azure consumer grade depth camera one can achieve a precision that is only few tens of millimeters off.

Having shown that self-calibration of robotic system is possible there are still multiple possible extensions to the formulated approach to be investigated:
As the system is already capable to undistort scans obtained from projections along a badly parameterized kinematic chain it should also be possible to include a sensor's intrinsic parameters in the optimization.
Also finding formulations to deal with less precise actuator readings such as odometry would allow to further grow the area of possible applications.
Finally, many strategies for optimizing the runtime of the ICP algorithm have been demonstrated.
It is not unlikely that many of those are applicable in the context of a calibration problem as well.

\section*{Appendix: Acknowledgment and Implementation Details}

Special thanks goes to Daniel Hettegger and Bare Luka \v{Z}agar for their continuous assistance in the lab and their remarks to the contents and structure of this work.
This also applies to Dinesh Paudel who was a great help in the assembly of the used robot workcell.

The presented C++ implementation would not have been possible without many contributors of open source libraries, most importantly Eigen \cite{eigenweb} and Ceres Solver \cite{ceres-solver}.
Also great thanks to Salvatore Virga for publishing his iiwa\_stack \cite{hennersperger2017towards}.

\ifCLASSOPTIONcaptionsoff
  \newpage
\fi
\balance

\vspace{7mm}%
\bibliographystyle{IEEEtran}
\bibliography{literature}

\end{document}